\documentclass[sn-basic]{sn-jnl}

\usepackage{easy-todo}
\usepackage{natbib}

\jyear{2022}%

\newif\ifarxiv
\arxivtrue
\begin{document}

% Define abbreviations and macros

\newcommand{\rmu}{LAM:ECHR}
\newcommand{\bert}{\textsf{BERT}}
\newcommand{\bertbs}{\textsf{BERT-Base}}
\newcommand{\lbert}{\textsf{Legal-BERT}}
\newcommand{\roba}{\textsf{RoBERTa}}
\newcommand{\robalg}{\textsf{RoBERTa-Large}}
\newcommand{\lrobalgt}{\textsf{Leg-RoBERTaL-13k}}
\newcommand{\lrobalgf}{\textsf{Leg-RoBERTaL-15k}}

\newcommand{\claw}{CaseLaw}
\newcommand{\jrc}{JRC-Acquis-En}

\title[Mining Legal Arguments in Court Decisions]{Mining Legal Arguments in Court Decisions}

\author*[1]{\fnm{Ivan} \sur{Habernal}}\email{ivan.habernal@tu-darmstadt.de}
\author[1]{\fnm{Daniel} \sur{Faber}}
\author[4]{\fnm{Nicola} \sur{Recchia}}
\author[2]{\fnm{Sebastian} \sur{Bretthauer}}
\author[3]{\fnm{Iryna} \sur{Gurevych}}
\author[2]{\fnm{Indra} \sur{Spiecker genannt Döhmann}}
\author[2]{\fnm{Christoph} \sur{Burchard}}

\affil*[1]{\orgdiv{Trustworthy Human Language Technologies}, \orgname{Technical University of Darmstadt}, \orgaddress{\country{Germany}}}
\affil[2]{\orgdiv{Faculty of Law}, \orgname{Goethe-Universität Frankfurt am Main}, \orgaddress{\country{Germany}}}
\affil[3]{\orgdiv{Ubiquitous Knowledge Processing (UKP) Lab}, \orgname{Technical University of Darmstadt}, \orgaddress{\country{Germany}}}
\affil[4]{\orgdiv{Department of Legal, Language, Interpreting and Translation Studies}, \orgname{University of Trieste}, \orgaddress{\country{Italy}}}

\abstract{
Identifying, classifying, and analyzing arguments in legal discourse has been a prominent area of research since the inception of the argument mining field. However, there has been a major discrepancy between the way natural language processing (NLP) researchers model and annotate arguments in court decisions and the way legal experts understand and analyze legal argumentation. While computational approaches typically simplify arguments into generic premises and claims, arguments in legal research usually exhibit a rich typology that is important for gaining insights into the particular case and applications of law in general.
We address this problem and make several substantial contributions to move the field forward. First, we design a new annotation scheme for legal arguments in proceedings of the European Court of Human Rights (ECHR) that is deeply rooted in the theory and practice of legal argumentation research. Second, we compile and annotate a large corpus of 373 court decisions (2.3M tokens and 15k annotated argument spans). Finally, we train an argument mining model that outperforms state-of-the-art models in the legal NLP domain and provide a thorough expert-based evaluation. All datasets and source codes are available under open lincenses at \url{https://github.com/trusthlt/mining-legal-arguments}.}


\ifarxiv
\thispagestyle{empty}

\noindent \textbf{Mining Legal Arguments in Court Decisions}

\medskip
\noindent Ivan Habernal, Daniel Faber, Nicola Recchia, Sebastian Bretthauer, Iryna Gurevych, Indra Spiecker genannt Döhmann, and Christoph Burchard

\bigskip
This is a \textbf{final pre-print version} of the article accepted for publication in \emph{Artificial Intelligence and Law}. The final official version will be published by Springer at \url{https://www.springer.com/journal/10506/}

\medskip
Submitted: August 5, 2022; Revised: May 5, 2023; Accepted: May 17, 2023

\medskip
Please cite this paper as follows.
\medskip

\begin{verbatim}
@article{Habernal.et.al.2023.AILaw,
  title   = {{Mining Legal Arguments in Court Decisions}},
  author  = {Habernal, Ivan and
  	         Faber, Daniel and
  	         Recchia, Nicola
  	         and Bretthauer, Sebastian and
  	         Gurevych, Iryna and
  	         Spiecker genannt D\"{o}hmann, Indra and
  	         Burchard, Christoph},
  year    = 2023,
  journal = {Artificial Intelligence and Law},
}
\end{verbatim}
\newpage
\fi

\maketitle

\section{Introduction}
\label{sec:introduction}

One of the most important tasks of law is to provide resolution mechanisms for conflicts. For this purpose, there are numerous laws which permeate almost all areas of daily life. If this leads to legal disputes between state institutions and citizens or between private individuals, courts must decide these conflicts and interpret them on the basis of the relevant legal norms. It is thus the primary task of the courts, especially in public law, to rule on the lawfulness of state decisions. Therefore, according to legal theory, argumentation forms the backbone of rational and objective decision-making in court proceedings.

Legal argumentation research spans also a wide variety of disciplines and a rich history dating back to the early days of `classical' scholars in the second century BC, with some modern twists by contemporary philosophers \citep{Toulmin.1958}. The end of the 20th century brought legal argumentation to the research agenda of AI \citep{Skalak.Rissland.1992}. Attempts to automatically identify, classify, and analyze arguments in legal cases stood at the beginning of early works in the field of `argument mining' \citep{Mochales.Moens.2008}.

To date, however, there has been a major discrepancy between the way legal experts analyze legal argumentation and the way natural language processing (NLP) researchers model, annotate, and mine legal arguments. While computational approaches typically treat arguments as structures of premises and claims \citep{Stede.Schneider.2018}, arguments in legal research usually exhibit a rich typology that is important for understanding how parties argue  \citep{Trachtman.2013}.

This paper aims to fill this gap by addressing the following research questions. First, we ask how reliably we can operationalize models of arguments from legal theory in terms of discourse annotations. Second, we want to explore how we can develop robust argument mining models that outperform the state of the art under the constraints of extremely expensive, expert-labeled data.

Our work makes several important contributions. First, we design a new annotation scheme for legal arguments in proceedings of the European Court of Human Rights (ECHR) that is deeply rooted in the theory and practice of legal argumentation research 
 \citep{Grabenwarter.Pabel.2021,Ruthers.et.al.2022,Schabas.2015,Ammann.2019,Barak.2012}.
Second, we compile and annotate a large corpus of 373 court decisions (2,395,100 tokens and 15,205 annotated argument spans) from the ECHR covering Articles 3, 7, and 8 of the European Convention on Human Rights. Third, we develop an argument mining model that outperforms state-of-the-art models in the legal NLP domain and provide a thorough expert-based evaluation. Finally, we preliminarily experiment with supervised linear models to investigate whether particular legal argument patterns affect the overall importance of the case as determined by the ECHR Bureau. All datasets and source codes are available at \url{https://github.com/trusthlt/mining-legal-arguments}.

\section{Related work}
\label{sec:related.work}

We review existing works in the legal domain, in particular argument mining in legal texts and works dealing with the ECHR judgments. Additionally, we focus on domain-specific pre-training that gives the background for our further experiments with language modeling.

\paragraph{Argument Mining in Court Decisions}

In their earliest work, \citet{Moens.et.al.2007} proposed a binary classification of isolated sentences as either argumentative or non-argumentative, on a corpus\footnote{Araucaria \citep{Reed.Gowe.2004}} containing five court reports from the UK, US and Canada, among others. However, no further details about the exact way of selecting the data, expertise of the annotators, or the agreement scores are known.

This work was later extended by \cite{Mochales.Moens.2007} who fine-grained the previously binary classification into three classes, namely premise, conclusion, or non-argumentative sentence. Moreover, each sentence was classified in the context of the preceding and the succeeding sentence. This work also introduced ECHR for the first time, where two lawyers annotated 12k sentences from 29 admissibility reports and 25 "legal cases" over the course of four weeks. This resulted in a corpus of roughly 12k sentences, with the majority (10k) non-arguments, 2,335 conclusions, and 419 premises. Despite being pioneering in analyzing ECHR, there are two open questions. The minor one is that the paper lacks any information on the annotator agreement. The major one is the very motivation for using this particular scheme of premise/conclusion/non-argument for analyzing legal argumentation. The authors developed their annotation scheme by taking an indirect inspiration from \citet{Walton.1996} and claim that this would \emph{``enable one to identify and evaluate common types of argumentation in everyday discourse"}. However, the referenced Walton's works \cite{Reed.Walton.2001} hardly deals with legal argumentation.\footnote{
Whether or not Walton's argument schemes are suitable for analyzing legal discourse in the first place lacks solid empirical evidence. \citet[p.~300]{Feteris.2017} surveys Walton's works and points out a manual analysis of the case \emph{Popov v. Hayashi} \citep{Walton.2012}. The analysis is, however, detached from the actual judgment text as the arguments are manually extracted, rephrased, and put into an argument diagram. This approach does not seem to have been widely adopted by others.
} The utility of this scheme for legal argument analysis thus remains unaddressed.

A similar motivation for using Walton's schemes and the premise/conclusion model has been later adopted in consecutive works by the same authors \citep{Mochales.Moens.2008,Mochales.Moens.2011} where the latter defines an argument as a set of propositions that adhere to one of the Walton's argumentation schemes and thus can be challenged by critical questions. \cite{Mochales.Moens.2008} studied 10 judgments and decisions from ECHR and obtained a Kappa agreement between two independent lawyers of 0.58. In a follow up, \cite{Mochales.Moens.2011} experimented with 47 annotated ECHR documents from which only \emph{The Law} section had been selected for annotation. While the authors original aim was to annotate a tree structure over the entire document, the actual corpus consisted of implicit relations between sentences, where a list of consecutive sentences is grouped into an argument, and each sentence is labeled either as support, against, or conclusion. It is unclear whether the Walton's schemes had been assigned to the arguments.

More recently, \cite{Poudyal.et.al.2020.ArgMinWS} published an annotated corpus of 42 ECHR decisions based on the previously annotated corpus by \cite{Mochales.Moens.2008}. They evaluated three tasks using a simple \roba\ model on their published dataset. First, clause detection identifies whether a clause belongs to an argument or not. Second, argument relation prediction for each pair of arguments decides whether they are related. Third, premise and conclusion recognition decides which clause is a premise and which is a conclusion. The last two tasks require perfect recognition of the clauses from the first task. The dataset is in a JSON format with pre-extracted sentences, such that the full original texts of judgments are not available.

With the goal of summarizing court decisions, \cite{Yamada.et.al.2019.AILaw} annotated 89 Japanese civil case judgments (37k sentences) with a tree-structured argument representation; a feature typical to these particular legal documents. They used four phases of annotations performed by a Japanese law Ph.D. student and reported experiments on classifying each sentence into one of the seven classes \citep{Yamada.et.al.2019.JURIX}.

\cite{Xu.et.al.2020.JURIX} explored argument mining to improve case summaries, which should contain the following key information: 1) the main issues the court addressed in the case, 2) the court's conclusion on each issue, and 3) a description of the reasons the court gave for its conclusion. They called these key pieces of information "legal argument triples" and intended to use them to create concise summaries, for which they annotated the human summaries with these triples, followed by annotating the text in the full court case corresponding to the summary triples. Finally, they performed argument mining to extract these triples from both the summaries and the full court case files.

A recently published task on legal argument reasoning \citep{Bongard.et.al.2022.NLLP} deals with exploring legal reasoning capabilities of language models, but does not address mining arguments from long text documents.

Compared to previous work, we depart from the usual premise-conclusion scheme by using a novel annotation scheme taken from a legal research perspective. Moreover, our dataset on the ECHR is much larger, with a total of 393 annotated decisions.

\paragraph{Pretraining large language models} \label{sec:rl_pretrain}

\cite{Gururangan.et.al.2020.ACL} examined the effects of continued pre-training of  language models such as \roba\ on domain- and task-specific data. To this end, they studied four domains, namely news, reviews, biomedical papers, and computer science papers. First, they pretrained a masking model on a large corpus of unlabeled domain-specific text, which they called domain adaptive pretraining. They found that this consistently improved performance on tasks in the target domain. Second, they investigated whether to continue pretraining only on the task-specific training data, which they called task-adaptive pretraining. They found that it consistently outperformed the \roba\ baseline and matched the performance of domain adaptive pretraining in some tasks. Finally, they combined the two procedures, running domain adaptive pretraining first and then task-adaptive pretraining, which performed best.

\cite{Gu.et.al.2022.ACM.CH} investigated pretraining in a specialized biomedical domain using biomedical abstracts from PubMed. They found that domain-specific pretraining from scratch worked best, along with the creation of a new domain-specific vocabulary for the pretraining process. In addition, masking whole words instead of masking sub-word tokens improved performance. They concluded that, as in the case of biomedicine, it may be more beneficial to pre-train from scratch if a sufficient amount of domain-specific data is available.

\cite{Chalkidis.et.al.2020.Findings} compared different approaches to adapting \bert\ to legal corpora and tasks by using vanilla \bert, adapting \bert\ by further pre-training on a domain-specific legal corpora, and pre-training \bert\ from scratch on a domain-specific legal corpora similar to the pre-training of the original \bert. They compared these approaches on classification and sequence labeling tasks and found that the best model can vary between further pre-training and pre-training from scratch depending on the task. They also found that the more challenging the final tasks, the more the model benefits from domain-internal knowledge. Finally, they published their model trained from scratch as \lbert.

\cite{Zheng.et.al.2021.ICAIL} examined the use of pretraining in the legal domain. They created a new dataset of Case Holdings on Legal Decisions with over 53,000 multiple-choice questions to identify the relevant statement of a cited case, a task they found to be legally significant and difficult from an NLP perspective. They pre-trained two different models. First, they continued the pre-training of \bert\ on their case law corpus. Second, they trained a model from scratch with a legal domain-specific vocabulary. They compared these two models to the standard \bert\ model as well as to the \bert\ model trained in twice the number of steps. They compared these models for different legal tasks and concluded that pretraining may not be worthwhile if the tasks are either too easy or not domain-specific in terms of the pretraining corpus. The more difficult and domain-specific the task, the greater the benefit of pretraining.

The above observations support our motivation for extensive pretraining. Since we want to extract and classify the arguments in a schema mainly used by lawyers, which we consider a very domain-specific task, we will also use the pretraining to build our own language model. 

\section{\rmu\ Corpus}

\subsection{Annotation scheme}
\label{sec:annotation.scheme}

This section introduces our new annotation scheme. From the NLP perspective, this scheme is 1) a text span annotation, 2) flat, non-hierarchical, and non-overlapping, 3) multi-class single-label, 4) aligned to tokens but independent of sentence boundaries, 5) cannot cross paragraphs as present in the court case and finally 6) each span is annotated with exactly two orthogonal tagsets, one for the argument type and one for the argument actor.

From the legal perspective the scheme is not based on logical categories of language (including legal language), as it is usual in the previously reviewed literature. Instead, we have chosen to break down the legal part of the ECHR's decisions by resorting to the usual legal categorization of the arguments used by ECHR. This allows for a deeper analysis of the Court's motivational itinerary, which is more similar to that of most jurists. Such an annotation, although much more complex in terms of NLP, has the advantage of enabling a fine-grained search that allows relevant arguments to be quickly found and filtered out.
For example, it is easy to identify in the Court's vast jurisprudence all the instances in which the Court uses a particular canon of argumentation, or to discuss, on the basis of empirical data, the quantitative or qualitative relationship in the use of different canons and also to observe their evolution over time.

Although the proposed annotation scheme has been tailored to ECHR decisions, the argument types we use are generally recognized in legal theory and can be found in many different courts. It should then be possible to apply the bulk of this annotation scheme to other courts, perhaps after some revision in the light of the particularities of the court in question. For example, the specifics of a country must be taken into account when interpreting constitutional documents. Further changes would be necessary for the Actors, where some of the entries would be deleted, such as Commission/Chamber, and others would certainly change their names.

The annotation scheme is divided into two main categories: the actors and their arguments. It should be noted that these two categories are completely orthogonal and independent, i.e., at least in theory, any actor can be linked to any kind of argument and vice versa.

\subsubsection{Actors} \label{sec:agent_types}

The annotation scheme covers five different types of actors.

\begin{enumerate}
\item \textbf{ECHR\ }
The ECHR is the most common agent and includes all arguments that the ECHR introduces.

\item \textbf{Applicant\ }
The applicant is the person or Contracting Party that litigates an alleged violation of a fundamental right as enshrined in the Convention.

\item \textbf{State\ }
The respondent means the party that is assumed by the applicant to be responsible for the alleged violation. Since the respondent is most of the time a Contracting Party, i.e. a state, we used this word for this category for the sake of clarity for the annotators.

\item \textbf{Third parties\ }
Third parties stand for all other parties that take part in the procedure, e.g., other Contracting Parties or NGOs such as Amnesty International or Human Rights Watch.

\item \textbf{Commission/Chamber\ }
Finally, Commission/Chamber concerns all arguments that originate from the Commission (until this organ was eliminated with the entry into force of Protocol no. 11 to the Convention in 1998) or from a Chamber, in the event of a Great Chamber decision, and are merely reproduced by the ECHR.
\end{enumerate}

\subsubsection{Argument Types} \label{sec:arg_types}

The annotation scheme includes sixteen different categories of arguments. Each of these categories is the subject of an in-depth and complex analysis in legal theory. Our aim to take them up in their canonical content and give a brief definition; this was already done at the beginning of our project for the benefit of the annotators (who were also given concrete examples from ECHR judgments in addition to such brief definitions, see Appendix \ref{appendix:arg_types}).

\begin{enumerate}
\item \textbf{Procedural arguments -- Non contestation by the parties\ }
This category describes the situation of consensus on a fact or argument between the parties, that allows the Court not to discuss further the matter, so that judicial time and resources can be saved.

\item \textbf{Method of interpretation -- Textual interpretation\ }
The textual interpretation, as can be found in Art.~31 §1--3 and in Art.~33 §1--3 of the Vienna Convention on the Law of Treaties (VCLT), is usually seen as the starting point for the interpretation of a norm. Textual interpretation can be referred to the meaning of the norm wording at the time of its origin or its application as well as its meaning in the technical or (most subsidiarily) colloquial language (\citeauthor{Ruthers.et.al.2022}, \citeyear{Ruthers.et.al.2022}, §731 et seq.;
\citeauthor{Ammann.2019}, \citeyear{Ammann.2019}, §197 et seq.). Also in the specific context of the ECHR, the textual interpretation and the corresponding norm of the VCLT have been seen as ``the backbone for the interpretation of the Convention'' (see, also more generally on the role of the VCLT in the case-law of the ECHR, \citeauthor{Schabas.2015}, \citeyear{Schabas.2015}, 34 et seq.).

\item \textbf{Method of interpretation -- Historical interpretation\ }
The historical interpretation as foreseen by Art. 31 para. 4 and Art. 32 VCLT involves the analysis of the historical circumstances at the moment of the enactment of the norm in order to ascertain its objective; cf.~further (\citeauthor{Ruthers.et.al.2022}, \citeyear{Ruthers.et.al.2022}, §778 et seq.;
\citeauthor{Ammann.2019}, \citeyear{Ammann.2019}, §778 et seq.). As foreseen by Art.~31 §4 and Art.~32 VCLT it is only of subsidiary importance in international treaties \citep[§5]{Grabenwarter.Pabel.2021}. This is especially true with reference to the ECHR since its preparatory work ``are rather sparse, especially with respect to the definitions of fundamental rights and issues related to their interpretation and application'' \citep[45~et~seq.]{Schabas.2015}.

\item \textbf{Method of interpretation -- Systematic interpretation\ }
The systematic interpretation -- as provided by in Art.~31 §3 lit. c VCLT -- is based on the ideal of a self-consistent legal system. Each norm is thus to be interpreted only from its position and function within the complete legal system (\citeauthor{Ruthers.et.al.2022}, \citeyear{Ruthers.et.al.2022}, §744 et seq.;
\citeauthor{Ammann.2019}, \citeyear{Ammann.2019}, p.~202). The ECHR has clearly recognized a duty to interpret the Convention in harmony with other rules of international law of which it forms part \cite[see further][37~et~seq.]{Schabas.2015}. At the same time, the Court also insists that “the Convention must be read as a whole, and interpreted in such a way as to promote internal consistency and harmony between its various provisions” \cite[see further][47]{Schabas.2015}.

\item \textbf{Method of interpretation -- Teleological interpretation\ }
We have encompassed in the same category a number of similar if by no means identical arguments to which the ECHR sometimes resorts. 
First of all, the teleological interpretation stricto sensu, as foreseen by Art.~33 §4 VCLT, i.e. the hermeneutic argument that is concerned with the objective (telos) that is to be achieved by the norm. The decisive factor is here not the historical intention of the legislator, but the objective purpose expressed in the norm which is characterized significantly by the textual, systematic and historical interpretation (\citeauthor{Ruthers.et.al.2022}, \citeyear{Ruthers.et.al.2022}, §717 et seq.;
\citeauthor{Ammann.2019}, \citeyear{Ammann.2019}, p.~208). Together with the teleological interpretation this category is also used for the dynamic or evolutive interpretation of the ECHR, which has been seen by the Court as a ``living instrument'' (\citeauthor{Grabenwarter.Pabel.2021}, \citeyear{Grabenwarter.Pabel.2021}, §5 para.~14 et seq.;
\citeauthor{Schabas.2015}, \citeyear{Schabas.2015}, 47~et~seq.).
This dynamic or evolutive interpretation of the ECHR is often also associated to the idea that the ECHR should offer an effective protection of fundamental rights (on the principle of effectiveness in the context of the ECHR see further \cite[49~et~seq.]{Schabas.2015}).

\item \textbf{Method of interpretation -- Comparative law\ }
With this argument type, all instances were annotated in which the ECHR makes references to legal provisions or case law of the Contracting Parties or to other legal orders, such as prominently the EU and the case law of the Court of Justice of the European Union. Very common are also references to international law more generally than in the stricter sense relevant in the sense of Art.~31 §3 lit.~c VCLT for the systematic interpretation \cite[see further][38~et~seq.]{Schabas.2015}.

\item \textbf{Test of the principle of proportionality -- Legal basis\ }
In a constitutional democracy, a constitutional right cannot be limited unless such a limitation is authorized by law. This is the principle of legality. From here stems the requirement -- which can be found in modern constitutions' limitation clauses, as well as in other international documents -- that any limitation on a right be ``prescribed by law''. At the basis of this requirement stands the principle of the rule of law
(\citeauthor{Barak.2012}, \citeyear{Barak.2012}, p.~107;
with specific reference to the ECHR context
\citeauthor{Grabenwarter.Pabel.2021}, \citeyear{Grabenwarter.Pabel.2021}, §18 para.~7 et seq.)

\item \textbf{Test of the principle of proportionality -- Legitimate purpose\ }
This component of the proportionality test examines whether it is possible to ascertain a legitimate purpose in the law that limits a fundamental right
(cf.\ further in general
\citeauthor{Barak.2012}, \citeyear{Barak.2012}, p.~107;
with specific reference to the ECHR context again
\citeauthor{Grabenwarter.Pabel.2021}, \citeyear{Grabenwarter.Pabel.2021} §18 para.~7 et seq.)

\item \textbf{Test of the principle of proportionality -- Suitability\ }
This component of the proportionality test analyses whether the means chosen by the law fit, i.e. can effectively realize or advance, the legitimate purpose of the law itself
(\citeauthor{Barak.2012}, \citeyear{Barak.2012}, p.~303 et seq.;
\citeauthor{Grabenwarter.Pabel.2021}, \citeyear{Grabenwarter.Pabel.2021}, §18 para.~15).

\item \textbf{Test of the principle of proportionality -- Necessity/Proportionality\ }
Since the ECHR does not strictly differentiate between the categories of necessity and proportionality in a strict sense
\citep[see][§18 para.~15]{Grabenwarter.Pabel.2021},
considerations of necessity -- if present -- also fall into this category. 
The test of necessity dictates that the legislator chooses -- of all suitable means -- only the ones that would limit at least the human right in question
\citep[see generally][p.~317 et seq.]{Barak.2012}.
On the other end the component of proportionality stricto sensu dictates that, ``in order to justify a limitation on a constitutional right, a proper relation (`proportional' in the narrow sense of the term) should exist between benefits gained by the public and harm caused to the constitutional right from obtaining that purpose''
(\citeauthor{Barak.2012}, \citeyear{Barak.2012}, p.~340 et seq.;
\citeauthor{Grabenwarter.Pabel.2021}, \citeyear{Grabenwarter.Pabel.2021}, §18 para.~14 et seq.)

\item \textbf{Institutional arguments -- Overruling\ }
This category refers to the amendment of a precedent on a horizontal level. It can only be done under the premise of fundamental deficits of the previous precedent. Since an overruling is extremely rare, it both fulfils the requirement of continuity and legal certainty and allows for viability and adaptability
(cf.~further \citeauthor{Maultzsch.2017}, \citeyear{Maultzsch.2017}, para.~1342;
for an in-depth analysis regarding the ECHR see \citeauthor{Mowbray.2009}, \citeyear{Mowbray.2009}).

\item \textbf{Institutional arguments -- Distinguishing\ }
This category is relevant when looking at a precedent and assessing an essential difference of facts, which leads to a non-transfer of a precedent to the new case
\citep[cf.~further][para.~1346]{Maultzsch.2017},

\item \textbf{Institutional arguments -- Margin of Appreciation\ }
The margin of appreciation is a margin of discretion granted by the ECHR to the judiciary, legislature and executive of the Member States before a violation of the ECHR is assumed and is functional to modulate the strictness of the review by the ECHR in different areas and contexts
(\citeauthor{Grabenwarter.Pabel.2021}, \citeyear{Grabenwarter.Pabel.2021}, §18 para.~20 et seq.;
\citeauthor{Schabas.2015}, \citeyear{Schabas.2015}, 78 et seq.)

\item \textbf{Precedents of the ECHR\ }
This category concerns the effect of the legal content of earlier judgments of the ECHR for later judgments
\cite[para.~1330 et seq.]{Maultzsch.2017}.
In the context of the ECHR there is not a rule of binding precedent or stare decisis, but nevertheless the Court normally follow its precedents
\cite[cf.\ further][46 et seq.]{Schabas.2015}.

\item \textbf{Decision of the ECHR\ }
This category contains the decisions of the ECHR which can be the final sentence on the result of the interpretation of a norm as well as the final sentence of the part of the judgment on the application to the concrete case.

\item \textbf{Application to the concrete case\ }
Application to the concrete case is concerned with determining the relation between the concrete case and the abstract legal norm by the subsumption of the facts of a case under a legal norm, i.e. examining whether the offence is fulfilled and the legal consequence thereby triggered
\citep[para. 677 et seq.]{Ruthers.et.al.2022}.

\end{enumerate}

\subsection{Data and annotation process} \label{sec:echr_data}

We scraped and extracted a large `raw' data collection from the HUDOC web interface\footnote{\url{https://hudoc.echr.coe.int/}} which contains, among others, judgements and decisions following a relatively rigid structure. A case begins with the list of the judges of the court, the registar, the indication of the applicant, the respondent and possibly other parties that have been admitted to the case. Next, the procedure before the Court and facts of the case are described. After the facts, in the "The Law" Section the arguments of the parties and of the Court on each alleged violation of the Convention are presented. Finally, the Courts renders its decision; see an example in Appendix~\ref{appendix:judgment}. All judgments that were later selected for annotation were processed by a bespoke html-to-xml extraction that retained the case and paragraph structure, and all paragraphs were tokenized using spaCy.\footnote{\url{https://spacy.io}}

\subsubsection{Annotation process} \label{sec:am_dataset}

We hired six law students for a 12-month period as annotators, supervised by two postdoctoral researchers (experts in public and criminal law) and consulted by two law professors (also public and criminal law).
We randomly selected ECHR cases concerning Article 8 (right to respect for private and family life, home and correspondence) and also to a lesser degree Article 7 (no punishment without law) to match our legal expertise. We also tried to balance cases in terms of their importance (from level one to four) as well as their recency.

The annotation process was conducted using the INCepTION platform \citep{Klie.et.al.2018}  and consisted of multiple rounds in which the annotators were given feedback about fully and partially missing annotations. In the end, we annotated 375 ECHR cases from which we excluded two documents as they had less than five arguments.

We measured pairwise annotation agreement using Krippendorff's unitized alpha $\alpha_u$ \citep{Krippendorff.2014} as implemented by \cite{Meyer.et.al.2014}. In the first annotation round, the $\alpha_u$ was in 0.70s. After the first round the causes of the disagreements were discussed and the annotation guidelines were updated to reflect that. In the rounds after that the $\alpha_u$ was in the 0.80s with a few rare outliers where the score degraded by around 0.2.\footnote{In particular, we measured the agreement only in \emph{The Law} section of the documents as the preceding sections were not annotated by design. The first six batches of annotations had the following $\alpha_u$ scores for type and actor, respectively: $\{0.70, 0.98, 0.96, 0.82, 0.84, 0.93\}$, $\{0.67, 0.97, 0.93, 0.80, 0.79, 0.90\}$. Each document in this batch was annotated by three annotators. We will release the full code for agreement computation in the project's GitHub repository.}
Values of $\alpha_u$ above 0.80 are considered very high for span annotations. This indicates that even for human experts the argument annotation is a difficult task. For the final single gold-standard dataset, the annotations were manually curated and merged by an independent expert annotator who solved disagreements. Initially the documents were annotated by all six annotators, then we ran several batches with three persons per document (those were used for calculating the inter-annotator agreement) and the last part of the study was ran by each annotator independently.

Our final gold-standard dataset \rmu\ consists of 373 ECHR cases annotated in the UIMA-XMI stand-off annotation format which keeps the original text along with span information about tokens, paragraphs and arguments as well as their labels. This enables easy post-processing by frameworks such as \texttt{DKPro} \citep{eckart-de-castilho-gurevych-2014-broad} or \texttt{dkpro-cassis} for any arbitrary downstream experiments. Furthermore, we also release the individual raw annotations from each annotator to facilitate future work for evaluation human judgments under uncertainty, such as in \citep{Simpson.Gurevych.2019}.

\subsubsection{Representing arguments for argument mining}

Since a single ECHR case is easily multiple thousand words long and the models we work with cannot handle such long sequences, we break down each case into smaller units. For this, we treat each \emph{paragraph} as a single short annotated text. This design choice is supported by the fact that legal arguments rarely span more than single paragraph in the ECHR judgments which was also considered in our annotation scheme (see Section~\ref{sec:annotation.scheme}).

Similarly to \cite{Poudyal.et.al.2020.ArgMinWS} we trim the cases and only include paragraphs from \emph{The Law} section onwards. This has the advantages that training is avoided on paragraphs without arguments which also rapidly shortens the training time. 
We represent the argument spans using BIO encoding, see Figure~\ref{fig:bio.example} for an example.

\begin{figure}
\begin{small}
\hrule \vspace{0.5em}
\begin{verbatim}
22            O                       O
.             O                       O
In            B-Application case      B-Applicant
the           I-Application case      I-Applicant
applicants    I-Application case      I-Applicant
'             I-Application case      I-Applicant
submission    I-Application case      I-Applicant
,             I-Application case      I-Applicant
the           I-Application case      I-Applicant
refusal       I-Application case      I-Applicant
to            I-Application case      I-Applicant
grant         I-Application case      I-Applicant
...           ...                     ...
home          I-Application case      I-Applicant
country       I-Application case      I-Applicant
.             I-Application case      I-Applicant
\end{verbatim}
\vspace{0.5em} \hrule 
\end{small}
\caption{\label{fig:bio.example} An excerpt from a tokenized paragraph (first column) with the corresponding BIO labels for argument type and actor (second and third column, respectively).}
\end{figure}

Finally, we split the 373 cases into an training/dev/test sets, resulting in 299 cases for training, 37 for development and 37 for testing. The data were stratified such that the distribution of labels remains balanced across splits and to ensure that low-support labels are included in all splits.

\subsubsection{Data Analysis}

Additionally, we collected statistics about our gold data in general and for the arguments on the argument level as well as on the BIO-tag level. The whole corpus argument high-level statistic can be seen in Table \ref{table:arg_l_stats}. Since \emph{Historical interpretation} is not present in our dataset, we removed this argument type from our further work. Furthermore, we can already see that our argument type distribution is highly imbalanced with \emph{Application to the concrete case} making up more than 50\% of the dataset and around half of our labels having a support of less than 1\%, going as low as two instances of \emph{Overruling} in the whole dataset. This can be a hindrance in the model learning. In fact, a more detailed statistics on the BIO-tag level in Table \ref{table:bio_arg_stats} (Appendix~\ref{sec:bio-tag-distribution}) reveals that only six out of 31 argument labels have a support higher than 1\%. 

\begin{table} 
	\centering 
	\begin{tabular}{lrr } 
		\toprule
		Argument Type & Frequency & $\approx$ in \% \\ 
		\midrule
		Application to the concrete case & 8,688 & 57.14 \\
		Precedents of the ECHR & 2,015 & 13.25 \\
		Test of the principle of proportionality -- Proportionality & 1,958 & 12.88  \\
		\vspace{0.5em} Decision of the ECHR & 1,236 & 8.13 \\
		Test of the principle of proportionality -- Legal basis & 513 & 3.37 \\
		Test of the principle of proportionality -- Legitimate purpose & 271 & 1.78 \\
		\vspace{0.5em} Non contestation by the parties & 264 & 1.74 \\
		Distinguishing & 88 & 0.58 \\
		Margin of Appreciation & 67 & 0.44 \\
		Teleological interpretation & 50 & 0.33 \\
		Test of the principle of proportionality -- Suitability & 25 & 0.16 \\
		Comparative law & 13 & 0.09 \\
		\vspace{0.5em}Textual interpretation & 10 & 0.07 \\
		Systematic interpretation & 5 & 0.03 \\
		Overruling & 2 & 0.01 \\
		\bottomrule
	\end{tabular} 
	\caption{Number of text spans associated with a particular argument type in the entire annotated corpus} 
	\label{table:arg_l_stats} 
\end{table}

The high-level statistic of the \emph{agent} dimension in Table \ref{table:ag_l_stats} show a less severe skeweness due to only five categories in total. Even though \emph{ECHR} makes up nearly $2/3$ of the arguments, the two least common agent types have at least a support greater than 1\%. At the BIO-tag level (Table \ref{table:bio_ag_stats} again) both \emph{Third parties} and \emph{Commission/Chamber} drop under 1\%. This can also be a challenge for model learning, although it might not be as bad as for the argument type because most agent labels are still seen at least a few thousand times.

\begin{table} 
	\centering 
	\begin{tabular}{ lrr } 
		\toprule
		Agent & Frequency & in \% \\ 
		\midrule
		ECHR & 9,950 & 65.44 \\
		Applicant & 2,471 & 16.25 \\
		State & 2,399 & 15.78 \\
		Third parties & 212 & 1.39 \\
		Commission/Chamber & 173 & 1.14 \\
		\bottomrule
	\end{tabular} 
	\caption{Number of text spans associated with a particular argument agent in the entire annotated corpus} 
	\label{table:ag_l_stats} 
\end{table}

\section{Experiments}

\subsection{Datasets}
\label{chapter:datasets}

Apart from the new annotated \rmu\ corpus which we use for training and evaluation, we also acquired a collection of large unlabeled corpora for self-supervised pre-training of large language models. Table~\ref{table:data_overview} shows a comprehensive summary of all datasets we use for pre-training and for supervised experiments; details of the unlabeled corpora follow.

\begin{table} [h]
	\centering 
	\begin{tabular}{ lrrl } 
		\toprule
		Dataset & Court cases & Size (GB) & Usage \\ 
		\midrule
		\rmu\ train & 299 & -- & Training the arg.\ mining model \\
		\rmu\ dev & 37 & -- & Optimizing the arg.\ mining model \\
		\vspace{0.5em} \rmu\ test & 37 & -- & Evaluating the arg.\ mining model \\
		ECHR raw & 65,908 & 1.1 & Additional pretraining/From scratch \\
		\jrc\ & 23,545 & 0.3 & Additional pretraining \\
		\claw\ & 4,938,129 & 43.2 & Pretraining from scratch \\
		\bottomrule
	\end{tabular} 
	\caption{Overview of all datasets used in the experiments.} 
	\label{table:data_overview} 
\end{table}

\begin{itemize}
\item \textbf{ECHR raw\ } We additionally scraped another 172,765 ECHR cases and converted them into plain text. After discarding non-English or corrupted cases we ended up with 65,908 documents with a total file size of 1.1 GB.

\item \textbf{\jrc\ Corpus\ } The JRC-Acquis corpus by \cite{Steinberger.et.al.2006.LREC} from the European Commission Joint Research Center (JRC), is a parallel corpus in over 20 languages containing European Union (EU) documents of mostly legal nature. \emph{EU Acquis Communautaire} is the French name for the body of common rights and obligations which bind together all Member States of the EU. The Acquis consists of \emph{the contents, principles and political objectives of  the  Treaties;  EU  legislation;  declarations  and  resolutions;  international  agreements;  acts  and  common  objectives} \citep{Steinberger.et.al.2006.LREC}. 
We included the English part of the JRC-Acquis corpus. Even though it is not focused on court cases, it still is one of a few corpora concerning EU-wide legal text, and thus we deemed it as a good fit for our European legal pretraining corpus. Our \jrc\ sub-corpus consists of 23,545 documents converted to plain text.
\item \textbf{CaseLaw Corpus\ }
We crawled 4,938,129 US court cases from the Caselaw Access Project.\footnote{\url{https://case.law/}} It includes all official, book-published cases from all state courts, federal courts, and territorial courts for all US states as well as American Samoa, Dakota Territory, Guam, Native American Courts, Navajo Nation, and the Northern Mariana Islands from 1658 until 2019. We ignored any meta-data and only extracted plain-text \emph{opinions}, which contain the body of each case. In contrast to the ECHR cases, the structure of an opinion is not as rigid and vary a lot as the underlying database spans over 300 years of documents with different purposes. Nevertheless, most cases usually still include a background about the facts and procedure, followed by a discussion with a decision in the end, which makes this data suitable for legal language model pre-training. We refer to this dataset simply as \claw.	
\end{itemize}

\subsection{Models}

\subsubsection{Multitask fine-tuning with transformers}

We experiment with two downstream tasks: 1) labeling text spans with the argument type; and 2) labeling text spans with the agent from which perspective the argument is written. Since both tasks are sequence tagging task, are carried out over the same input data, which means in the same domain, and are related in the sense that an agent prediction (e.g. \textit{ECHR}) can help predicting the argument type (e.g. \textit{Decision of the ECHR}) or vice versa, we employ a multitask model.

We extend the huggingface transformers library \citep{Wolf.et.al.2020.EMNLP} and implement support for multitask fine-tuning.
In particular, we are instantiating two models, one with the \textit{ArgType} head, responsible for predicting the argument type, and one with the \textit{Agent} head, responsible for predicting the agent label. The heads share the same underlying encoder. This adaption allows us to easily instantiate our model while relying on existing model implementations from huggingface transformers, such as \bert. Furthermore, for each input batch we carry the additional information of the corresponding task along, such that we can dynamically map to the right model during training.

\subsubsection{Postprocessing} \label{sec:postprcess}

As transformer models operate with sub-word units, we map predictions back to the word level.\footnote{We created a special tag \texttt{-100} for all tokens starding with \texttt{\#\#
}.} This ensures a one-to-one correspondence in granularity of the predicted labels and the gold standard data labels, as both are then aligned on tokens (see Table~\ref{tab:mapping}). We thus also evaluate the models on the word level.

\begin{table}
\resizebox{\textwidth}{!}{
\begin{tabular}{l|lllllllll}
	\toprule
	Tokens & \multicolumn{3}{l}{entailed} &  by & Mr & . & \multicolumn{3}{l}{Berrehab} \\
	\midrule
	Subwords & en & \#\#tail & \#\#ed & by & Mr & . & Be & \#\#rre & \#\#hab \\
	Predictions & I-Sub & -100 & -100 & I-Sub & I-Sub & I-Sub & I-Sub & -100 & -100 \\
	Mapping & \multicolumn{3}{l}{I-Sub} & I-Sub & I-Sub & I-Sub & \multicolumn{3}{l}{I-Sub} \\
	\bottomrule
\end{tabular}
}
\caption{\label{tab:mapping} Example of sub-word level tag mapping to tokens to match the gold label BIO-tags granularity.}	
\end{table}

\subsubsection{Baseline models}

\paragraph{\roba}

\citet{Liu.et.al.2019.arXiv.RoBERTa} investigated hyperparameters in the pretraining of \bert\ and released an improved model \roba\ (Robustly optimized BERT approach). \roba\ modifies \bert\ in four ways: 1) the model is trained longer, on more data, and with bigger batch sizes; 2) it forgoes the next sentence prediction objective; 3) it is trained on longer sequences; and 4) the masking pattern is dynamically changed and applied during training. Additionally, \roba\ uses byte-level Byte-Pair Encoding \citep{Radford.et.al.2019.report}.

As our first baseline model we use \robalg, which has 24 layers, a hidden size of 1024 and 16 attention heads.
Our \roba\ model is fine-tuned for 10 epochs with a learning rate of $1e^{-5}$, batch size of $4$, weight decay of $0.01$ and $1000$ warmup steps.

\paragraph{\lbert}

As a second baseline, we use the \lbert\ model by \cite{Chalkidis.et.al.2020.Findings} which performed better on legal tasks than the corresponding \bert\ model. They pretrained \lbert\ with the same configuration of \bertbs\ from scratch, using their own newly created vocabulary. As the data for the pretraining, they collected nearly 12 GB of data from the legal sub-domains legislation, cases and contracts.

Our \lbert\ model, like our \roba\ model, is fine-tuned for 10 epochs with a with a learning rate of $1e^{-5}$, batch size of $8$, weight decay of $0.01$ and $500$ warmup steps.

\subsection{Pre-training for robust domain adaptation} \label{sec:om}

We investigate to which extent further model adaptation helps to gain better downstream performance on legal argument mining in \rmu. As a main method, we further pretrain a baseline model on European legal data. This is motivated by the previous successes in domain adaptation discussed in Section~\ref{sec:rl_pretrain} \citep{Gururangan.et.al.2020.ACL,Gu.et.al.2022.ACM.CH,Chalkidis.et.al.2020.Findings,Zheng.et.al.2021.ICAIL,Yin.Habernal.2022.NLLP}. We also experimented with pretraining a language model from scratch on a big dataset of legal cases.\footnote{As this experiment failed, we report it only in Appendix~\ref{appendix:scratch}.}

\subsubsection{Further Pretraining}

As the basemodel for our further pretraining, we use the \robalg\ model because it performs slightly better than the \lbert\ model, but is not tuned to legal data in any way. For this reason, we expect a significant improvement when continuing the pretraining on legal data. 

As our pretraining data we use English ECHR cases and the \jrc\ dataset. The ECHR corpus has the size of 1.1 GB, the \jrc\ one of 338 MB, totaling to around 1.4 GB. The ECHR corpus contains 65,908 cases, the JRC-Aquis of 23,545 legal documents.

We pre-train our models with masked language modeling. We further pretrain our model for 15,000 steps (around 51 epochs) which is a middle ground between the domain-adaptive and task-adaptive pretraining  of \citet{Gururangan.et.al.2020.ACL}. Similarly, we use a batch size 2048 and a learning rate of 5e-4 with a warmup ratio of 6\%. 
The loss generally decreases over time, from initially 0.9567 to 0.4435 at step 15,000. Similarly, the loss in our validation data started at 0.6106 decreased to 0.4310.

\subsection{Downstream task results and analysis}

For each of our models, we selected the fine-tuned model checkpoint with the best combined performance on the dev set. These are after the 6th epoch for \lbert, after the 7th epoch for \robalg, after the 6th epoch for our further pretrained \robalg\ model for 13k steps on legal data (\lrobalgt) and after the 9th epoch for our further pretrained \robalg\ model for 15k steps on legal data (\lrobalgf). The Macro $F_1$ scores on the test and development data are in Table \ref{table:macro_f1}. 

The performance of both baseline models, \lbert\ and \robalg, are similar on the dev set.
On the test set, \robalg\ performs nearly two percentage points better contrary to our expectation as \lbert\ was pre-trained on legal data. However, the model size may play a role, too. \robalg\ is a larger model with 355 million parameters, compared to 110 million of \lbert. Additionally, \robalg\ was trained for 16 times more steps than \lbert\ when accounted for the batch size. This comparable performance in the development process also led us to use \robalg\ as the base for our further pretraining.

Our two models (\lrobalgt\ and \lrobalgf) show similar performance on the dev set, whereas on the test set the longer pretraining also led to a higher performance. They also perform significantly better compared to the other two baseline models, with the 2000 extra steps in our domain adaption leading to an improvement of nearly two percentage points on the test set. These improvements fall in line with the findings of \citet{Gururangan.et.al.2020.ACL} that continued pretraining and domain adaption generally lead to better results.

\begin{table} 
	\centering 
	\begin{tabular}{ lrrrr } 
		\toprule
		\multirow{2}{*}{Model} & \multicolumn{2}{c}{Argument Type} & \multicolumn{2}{c}{Agent} \\
		\cline{2-5}
		& Dev & Test & Dev & Test \\ 
		\midrule
		$\lbert$ \citep{Chalkidis.et.al.2020.Findings}& 39.42 & 37.38 & 84.82 & \textbf{91.79} \\
		\vspace{0.5em} $\robalg$ \citep{Liu.et.al.2019.arXiv.RoBERTa} & 40.10 & 39.14 & 86.30 & 90.83 \\
		\lrobalgt\ & \textbf{42.95} & 41.30 & \textbf{88.44} & 91.55 \\
		\lrobalgf\ & 42.81 & \textbf{43.13} & 87.68 & 91.36 \\
		\bottomrule
	\end{tabular} 
	\caption{Macro F$_1$-scores} 
	\label{table:macro_f1} 
\end{table}

\begin{table} 
	\centering 
	\begin{tabular}{lr|p{1.0cm}p{1.2cm}p{1.3cm}p{1.2cm}}
		\toprule
		Label & Freq. & \lbert\ & \robalg\ & \lrobalgt\ & \lrobalgf\ \\
		\midrule
		O & 83786  & 95.73 & 95.11 & 95.38 & \textbf{96.04} \\
		I-Application case & 65715 & 80.87 & 78.54 & 80.48 & \textbf{81.66} \\
		I-Precedents ECHR & 27515 & \textbf{86.32} & 84.61 & 84.99 & 84.50 \\
		I-Necessity/Proportionality & 20864 & 53.15 & 55.79 & 53.75 & \textbf{56.60} \\
		I-Legal basis & 6818 & 36.75 & 46.60 & 38.44 & \textbf{49.60} \\
		I-Decision ECHR & 3645 & \textbf{77.99} & 74.22 & 77.91 & 77.76 \\
		I-Distinguishing & 1805 & 42.66 & 47.70 & 45.34 & \textbf{53.44} \\
		I-Legitimate purpose & 996 & \textbf{86.18} & 74.30 & 67.24 & 73.67 \\
		I-Non contestation & 940 & \textbf{84.20} & 80.29 & 79.43 & 77.46 \\
		I-Teleological interpretation & 835 & 00.00 & 00.48 & 00.00 & \textbf{20.88} \\
		B-Application case & 801 & 80.68 & 78.95 & \textbf{81.32} & 80.77 \\
		I-Margin of Appreciation & 718 & 45.56 & 31.93 & 39.17 & \textbf{55.82} \\
		B-Precedents ECHR & 170 & 79.01 & \textbf{80.94} & 78.98 & 80.35 \\
		B-Necessity/Proportionality & 169 & 46.44 & \textbf{51.52} & 49.11 & 50.88 \\
		B-Decision ECHR & 129 & 71.65 & 70.72 & \textbf{72.65} & 71.02 \\
		I-Comparative law & 97 & 00.00 & 00.00 & \textbf{87.21} & 20.76 \\
		B-Legal basis & 65 & 38.64 & 46.85 & 41.82 & \textbf{47.06} \\
		I-Systematic interpretation & 40 & 00.00 & 00.00 & 00.00 & 00.00 \\
		I-Overruling & 39 & 00.00 & 00.00 & 00.00 & 00.00 \\
		B-Non contestation & 28 & 79.25 & \textbf{83.02} & 79.17 & 77.97 \\
		I-Textual interpretation & 23 & 00.00 & 00.00 & 00.00 & 00.00 \\
		I-Suitability & 19 & 00.00 & 00.00 & 00.00 & 00.00 \\
		B-Legitimate purpose & 18 & \textbf{73.68} & 71.79 & 63.16 & 61.90 \\
		B-Distinguishing & 16 & 00.00 & 20.00 & 27.27 & \textbf{38.71} \\
		B-Teleological interpretation & 12 & 00.00 & 00.00 & 00.00 & \textbf{13.33} \\
		B-Margin of Appreciation & 12 & 00.00 & 40.00 & 37.50 & \textbf{66.67} \\
		B-Comparative law & 2 & 00.00 & 00.00 & 00.00 & 00.00 \\
		B-Overruling & 1 & 00.00 & 00.00 & 00.00 & 00.00 \\
		B-Suitability & 1 & 00.00 & 00.00 & 00.00 & 00.00 \\
		B-Textual interpretation & 1 & 00.00 & 00.00 & 00.00 & 00.00 \\
		B-Systematic interpretation & 1 & 00.00 & 00.00 & 00.00 & 00.00 \\
		\bottomrule
	\end{tabular} 
	\caption{Test Macro F$_1$-scores of each argument type label sorted by frequency} 
	\label{table:arg_f1s} 
\end{table}

Table \ref{table:arg_f1s} breaks down the test set F$_1$-scores to the individual argument type labels. Generally, the more data exists for a class, the better the predictions for that class are. Exceptions to this pattern are two classes of the \textit{Test of the principle of proportionality}, namely \textit{Necessity/Proportionality} and \textit{Legal basis}, as well as the institutional argument \textit{Distinguishing}. We will further analyze these cases later. Our model \lrobalgt\ have the best F$_1$-scores in half of the labels. Additionally, it is the only model that has at least some correct predictions of the type \textit{Teleological interpretation}. Despite performing significantly worse on the Macro F$_1$-score, \lbert\ surprisingly still outperforms every other model on five labels, even significantly on \textit{Legitimate purpose} and \textit{Non contestation}, implying that it is hard for a single model to focus on all of the 15 categories and perform really well in all of them.

\begin{table} 
	\centering 
	\begin{tabular}{lr|p{1.1cm}p{1.4cm}p{1.4cm}p{1.4cm}}
		\toprule
		Label & Freq. & \lbert\ & \robalg\ & \lrobalgt\ & \lrobalgf\ \\
		\midrule
		I-ECHR & 90,263 & 96.55 & 96.24 & 96.43 & \textbf{97.06} \\
		O & 83,786 & 95.72 & 95.03 & 95.33 & \textbf{96.03} \\
		I-State & 19,392 & \textbf{94.95} & 94.48 & 93.61 & 94.61 \\
		I-Applicant & 15,526 & 92.52 & 91.25 & \textbf{92.93} & 92.83 \\
		I-Third parties & 2,962 & \textbf{95.76} & 92.53 & 91.99 & 92.11 \\
		I-Commission/Chamb. & 1,926 & 91.51 & \textbf{98.09} & 96.43 & 91.23 \\
		B-ECHR & 915 & 89.56 & 89.93 & 90.51 & \textbf{91.01} \\
		B-Applicant & 238 & \textbf{90.11} & 87.21 & 88.74 & 88.79 \\
		B-State & 219 & \textbf{88.13} & 87.67 & 87.10 & 87.42 \\
		B-Third parties & 28 & \textbf{90.91} & 88.89 & 88.89 & 89.29 \\
		B-Commission/Chamb. & 26 & 84.00 & 77.78 & \textbf{85.11} & 84.62 \\
		\bottomrule
	\end{tabular} 
	\caption{Test F$_1$-scores of each agent type label} 
	\label{table:ag_f1s} 
\end{table}

Regarding the agent predictions, the models vary by a few percentage points on the development set and are within one percentage point on the test set (see Table \ref{table:macro_f1}). All models perform strong on the agent prediction with F$_1$-scores over 90. Here the \lbert\ model also performs slightly better than the other models. Looking at the scores per label in Table \ref{table:ag_f1s}, we can also see that generally, the performance is better the more data for a class exists.

\subsubsection{Error analysis}

We analyze prediction errors both quantitatively and qualitatively. Since the agent is only of subsidiary importance and the model reaches a reasonable performance (91.36 Macro F$_1$), we only briefly discuss the most common classification error, namely the outside label predicted as \textit{I-ECHR}. These mistakes have in common that their text always include the court which probably leads to the misclassification, e.g., they include \emph{``the court considers that ...''} or \emph{``the court notes that ...''}.

For the argument types, we mainly examined labels with relatively high support but mediocre performance, namely \textit{Necessity/Proportionality}, \textit{Legal basis} and \textit{Distinguishing}. For this task, the legal experts from our team investigated individual instances of common errors and assessed the differences. A relevant remark is that for labeling arguments, legal experts often rely on the context beyond a single paragraph we used as our model input.

\begin{figure}[h!]
	\centering
	\includegraphics[width=\textwidth]{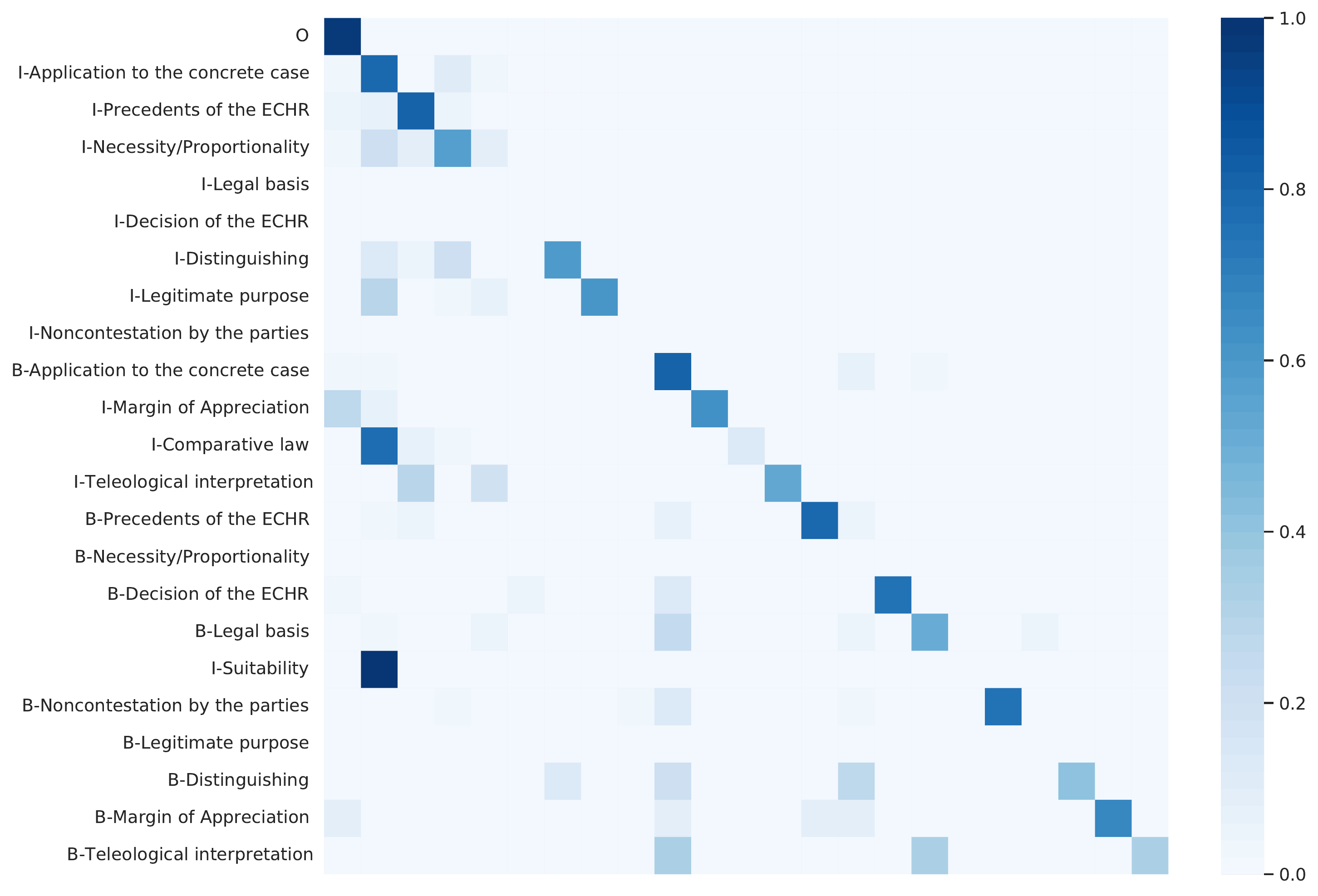}
	\caption{An excerpt from the probabilistic confusion matrix. Each row represents the gold label, each column the predicted label. Each cell value is the number of "row predicted as column" normalized per row. Rows are sorted in a decreasing frequency of examples, i.e.\ "O" ist the most common label and "B-Teleological interpretation" the least common; columns have the same ordering. Labels with zero predictions are omitted.}
	\label{fig:conf-matrix}
\end{figure}

Figure \ref{fig:conf-matrix} depicts the probabilistic confusion matrix in which each row is normalized to sum up to $1.0$.
Looking at the errors for \textit{Necessity/Proportionality}, by far the most common error is the misclassification as \textit{Application to the concrete case} in 6,992 instances. This category is also often misclassified as \textit{Precedents of the ECHR}. Other classes are also often predicted as \textit{Precedents of the ECHR}, especially as the second most common error \textit{Application to the concrete case} in 4,409 instances as well as \textit{Legal basis} (1,795 instances).

\paragraph{Examination by legal experts}

Detailed expert examination revealed that many of the mistakes of \textit{Necessity/Proportionality} as \textit{Application to the concrete case} and vice versa are not differentiable between these categories without more context. We further observed that without the context the predictions of our model are more convincing than the gold standard annotations.
For instance, in mistakes regarding \textit{Legal basis}, without the context we would often regard it as \textit{Application to the concrete case} instead of \textit{Legal basis} and \textit{Necessity/Proportionality}.

Detailed analysis of \textit{Necessity/Proportionality} misclassified as \textit{Precedents of the ECHR} revealed that our model technically recognized the text correctly as precedent but the nuances in the annotation guidelines defined that in \textit{Necessity/Proportionality} and \textit{Application to the concrete case} occurrences of \textit{Precedents of the ECHR} are not marked as such. Only little differences exist when considering the argument text here because in both cases the text contains references of the form \textit{``see X, § XX''}, e.g. Example \ref{ex:angemessenheit} \emph{``... (see Mamchur v. Ukraine, no. 10383/09, §100, 16 July 2015)''} was predicted as \textit{Precedents of the ECHR} despite being annotated as \textit{Necessity/Proportionality}. We suspect that differences in the other direction also stem from these seemingly inconsistencies that confuse our model, e.g. Example \ref{ex:precedent} seems really close to the previous example and even uses the phrase \emph{``fair balance''} which is often used for \textit{Necessity/Proportionality} and was predicted as such but is annotated as \textit{Precedents of the ECHR}.

\begin{figure}
	\begin{footnotesize}
		\textbf{Necessity/Proportionality:} 
		76 . In identifying the child ’s best interests in a particular case , two considerations must be taken into account : first , it is in the child ’s best interests that his or her ties with the family be maintained , except in cases where the family has proved particularly unfit or is clearly dysfunctional ; and second , it is in the child ’s best interests to ensure his or her development in a safe , secure and stable environment and in an environment which is not dysfunctional ( see Mamchur v. Ukraine , no . 10383/09 , § 100 , 16 July 2015 ) .
		\caption{\label{ex:angemessenheit} Argument misclassified as \emph{Precedents of the ECHR}}
	\end{footnotesize}
\end{figure}

\begin{figure}
	\begin{footnotesize}
		\textbf{Precedents ECHR:} 
		95 . The decisive issue is whether the fair balance that must exist between the competing interests at stake – those of the child , of the two parents , and of public order – has been struck , within the margin of appreciation afforded to States in such matters ( see Maumousseau and Washington , cited above , § 62 ) , taking into account , however , that the best interests of the child must be of primary consideration and that the objectives of prevention and immediate return correspond to a specific conception of “ the best interests of the child ” ( see paragraph 35 above ) .
		\caption{\label{ex:precedent} Argument misclassified as \emph{Necessity/Proportionality:}}
	\end{footnotesize}
\end{figure}

We envision several directions of future work to overcome these shortcomings. First, argument mining models could take the entire court decision document into account and model jointly all paragraphs at once; this it however not trivial with the majority of the current transformer models. Second, some adjustments in the annotation guidelines could solve the very ambiguous cases described above.

Finally, another line of future research could benefit from Bayesian treatment where we might directly train an ensemble model using all annotators' data instead of the curated gold standard, utilizing Bayesian sequence combination \citep{Simpson.Gurevych.2019}. This approach has been successfully tested on sequence labeling datasets with only a few labeled instances \citep{Simpson.et.al.2020.AAAI}. To which extent this method will generalize to larger tag-sets (the case of \rmu) remains an open research question.

\subsection{Generalization to Article 3} \label{sec:art3}

Additionally, we are interested in the performance of our model on cases concerning different Articles of the European Convention on Human Rights. We tested our best model \lrobalgf\ on an additional set of 20 annotated cases dealing with Article 3 (prohibition of torture). As our previous experiments mainly concerned Article 7 and Article 8, we expect worse results because each case starts from its normative text which is much more straightforward in Article 3 (\emph{No one shall be subjected to torture or to inhuman or degrading treatment or punishment}) compared to Article 8 (\emph{1. Everyone  has  the  right  to  respect  for  his  private  and  family life, his home and his correspondence ...}) and concerns a completely different topic. The results can be seen in Table \ref{table:art3_f1s}. 

\begin{table} 
	\centering 
	\begin{tabular}{lrrrr}
		\toprule
		Label & Frequency &  Original & Art.~3 & Difference (\%) \\
		\midrule
		Macro F1 &    183,878 &        43.13 &   31.49 &  --27 \\
		\midrule
		I-Application case &    104,344 &      81.66 &   90.02 &  10 \\
		
		O &     40,910 &      96.04 &   90.89 &  --5 \\
		
		I-Precedents ECHR &     32,571 &      84.50 &   83.98 &  --1 \\
		
		I-Decision ECHR &      2,911 &       77.76 &   79.84 &  3 \\
		
		\vspace{0.5em} B-Application case &       888 &       80.77 &   87.25 &  8 \\
		
		I-Distinguishing &       688 &       53.44 &   51.21 &  --4 \\
		
		I-Non contestation &       647 &       77.46 &   52.39 &  --32 \\
		
		I-Teleological Interpretation &       215 &       20.88 &    0.00 &  --100 \\
		
		B-Precedents ECHR &       199 &       80.35 &   80.00 &  0 \\
		
		\vspace{0.5em} I-Legal basis &       151 &       49.69 &   25.57 &  --49 \\
		
		B-Decision ECHR &        96 &       71.02 &   78.22 &  10 \\
		
		I-Necessity/Proportionality &        93 &       56.60 &    0.09 &  --100 \\
		
		I-Legitimate purpose &        59 &       73.67 &   67.84 &  --8 \\
		
		I-Textual interpretation &        39 &       0.00 &    0.00 &   0 \\
		
		\vspace{0.5em} I-Comparative law &        25 &       20.76 &    0.00 &  --100 \\
		
		I-Margin of Appreciation &        15 &       55.82 &    0.00 &  --100 \\
		
		B-Non contestation &        11 &       77.97 &   56.00 &  --28 \\
		
		B-Distinguishing &         6 &          38.71 &   37.50 &  --3 \\
		
		B-Necessity/Proportionality &         2 &          50.88 &    0.00 &  --100 \\
		
		\vspace{0.5em} B-Teleological Interpretation &         2 &          13.33 &    0.00 &  --100 \\
		
		B-Legal basis &         2 &          47.06 &   28.57 &  --39 \\
		
		B-Legitimate purpose &         1 &          61.90 &   66.67 &  8 \\
		
		B-Textual interpretation &         1 &          0.00 &    0.00 &   0 \\
		
		B-Comparative law &         1 &          0.00 &    0.00 &   0 \\
		
		B-Margin of Appreciation &         1 &          66.67 &    0.00 &  --100 \\
		\bottomrule
	\end{tabular} 
	\caption{Argument type F$_1$-scores on the Article 3 cases and comparison to the best scores on the original data (test dataset as shown in Table \ref{table:arg_f1s})} 
	\label{table:art3_f1s} 
\end{table}

The Macro F$_1$-score is only 31.49 compared to 43.13 on the original test data. Six classes are missing in the Article 3 data, namely the B and I tags of \textit{Systematic interpretation}, \textit{Suitability} and \textit{Overruling}. This also effects the Macro F$_1$-score but has no effect on the comparison because in the original data the F$_1$-scores of all these are 0. On the Article 3 data, the model only performs better for \textit{Application to the concrete case} and \textit{Decision of the ECHR} but fails completely on predicting \textit{Test of principle of proportionality -- Necessity/Proportionality} and \textit{Institutional arguments -- Margin of Appreciation}. 

Although not as large as before, the performance also degraded in the case of the agents (Table \ref{table:art3_ag_f1s}) with a Macro F$_1$-score of 83.15 compared to the original score of 91.36. Here, our model performs worse on every label on the Article 3 data. 

\begin{table} 
	\centering 
	\begin{tabular}{lrrrr}
		\toprule
		Label & Frequency &  Original & Art.~3 & Difference (\%) \\
		\midrule
		Macro F1 &    183,878 &        91.36 &   83.15 &  --9 \\
		\midrule
		I-ECHR &    100,784 &      97.06 &   94.51 &  --3 \\
		O &     40,910 &      96.03 &   91.10 & --5  \\
		I-Applicant &     20,018 &      92.83 &   85.70 &  --8 \\
		I-State &     14,819 &       94.61 &   90.13 &  --5 \\
		I-Third parties &      4,832 &       92.11 &   81.03 &  --12 \\
		I-Commission/Chamber &      1,305 &       91.23 &   61.26 &  --33 \\
		B-ECHR &       798 &       91.01 &   89.98 & --1  \\
		B-Applicant &       213 &       88.79 &   85.31 &  --4 \\
		B-State &       147 &       87.42 &   85.17 & --3  \\
		B-Third parties &        36 &       89.29 &   77.33 & --13  \\
		B-Commission/Chamber &        16 &       84.62 &   73.17 & --14  \\
		\bottomrule
	\end{tabular} 
	\caption{Agent F$_1$-scores on the Article 3 cases and comparison to the best scores on the original data (test dataset as shown in Table \ref{table:ag_f1s})} 
	\label{table:art3_ag_f1s} 
\end{table}

Overall, the transfer of the model to another Article performs significantly worse. This might be because cases of different Articles differ significantly in regard to the topic and normative text they cover, for example the \textit{Necessity/Proportionality} only occurs 93 times, whereas in the original data it makes up a much larger fraction with 20,864 occurrences, similarly \textit{Legal basis} with 151 vs. 6,818 occurrences.

\subsubsection{Limitations}
First, given the amount of computational resources needed for running all the above experiments, we had to restrain to testing only a single random seed. We acknowledge that having at least three to five runs and reporting an average and its standard deviation would give more robust performance estimates. We leave this for future work.

Second, the choice of BIO encoding, and in particular of the subsequent token-level evaluation, is overly pessimistic. Our metric penalizes a mismatch in argument component boundaries even if there might be a partial span match (the type of argument is recognized correctly). We refer here to existing literature on argument mining, namely \cite{Habernal.Gurevych.2017.CoLi}, who consider several alternative evaluation metrics but such a decision remains inconclusive without a specific down-stream task that might or might not cope with partially recognized arguments.

\subsection{Case study: Predicting case importance using arguments as features}

One of the potential use-cases for a collection of ECHR judgments with manually or automatically analyzed arguments is to perform an empirical analysis with respect to the importance of the case. We asked whether there is a potential link between argumentation pattern and the importance of the case as determined by the ECHR.

Each ECHR case has an importance level ranging from one to four, where level 1 is of the most and level 4 of the least importance. They correspond to the following schema:

\begin{description}
	\item[Level 1: Case Reports.] ``Judgments, decisions and advisory opinions delivered since the inception of the new Court in 1998 which have been published or selected for publication in the Court's official Reports of Judgments and Decisions.''\footnote{\url{https://www.echr.coe.int/Documents/HUDOC\_FAQ\_ENG.pdf}} These are the key cases of the most importance.
	\item[Level 2: High Importance.] All judgments, decisions and advisory opinions which make a significant contribution to the development, clarification or modification of its case-law, either generally or in relation to a particular State but are not included in the Case Reports.
	\item[Level 3: Medium Importance.] Other judgements, decisions and advisory opinions that, even though they are not making a significant contribution, still go beyond merely applying existing case law. 
	\item[Level 4: Low Importance.] ``Judgments, decisions and advisory opinions of little legal interest, namely  judgments  and  decisions  that  simply  apply  existing  case law,  friendly settlements and strike outs (unless raising a particular point of interest).''\footnotemark[\value{footnote}]
\end{description}

\subsubsection{Model and features} \label{sec:imp_model}

We employ a linear machine learning model, Support Vector Machines (SVMs), which allows us to inspect the prediction power of individual features. The features are listed in Table~\ref{tab:features}.
Since the performance of the SVM highly depends on the used hyperparameters, we employ a grid search.\footnote{
Kernels = linear, polynomial; $C \in \{0.1, 1, 10, 100, 1000\}$; $\gamma \in \{1, 0.1, 0.01, 0.001, 0.0001\}$; degree $\in \{2, 3, 4, 5, 6\}$
}

\begin{table}
\begin{tabular}{lp{4cm}p{5.7cm}}
\toprule
& Feature & Description \\
\midrule
1 & Doc.\ length & The length in tokens of the full case document \\
2 & Shortened doc.\ length & The length in tokens of the shortened document from \emph{AS TO THE LAW}/\emph{THE LAW} onwards (see end of Section \ref{sec:am_dataset}). \\
3 & Number of arguments & The total number of arguments in a case. \\
4 & Average argument length & The average length in characters of the arguments in a case. \\
5 & Fraction of argumentative part & The \textit{number of tokens in all arguments $/$ number of all tokens in the whole case document}. \\
6 & Shortened fraction of argumentative part & The \textit{number of tokens in all arguments $/$ number of all tokens in the shortened case document}. \\
7--21 & Fraction of argument $X$ & For each argument type the \textit{number of arguments of that type $X$ in a document $/$ number of all arguments in that case document}. \\
21--25 & Fraction of agent $X$ & For each agent the \textit{number of arguments with the agent $X$ in a document $/$ number of all arguments in that case document}. \\
\bottomrule
\end{tabular}	
\caption{\label{tab:features} Features for supervised importance prediction.}
\end{table}

The data is split in 80\% for training and 20\% for testing. By far the most common are cases of low importance (level 4), followed by those of medium importance (level 3), with the key cases of most importance (level 1) close behind and then the cases of high importance (level 2).

\subsubsection{Analysis}

The best performance in our grid search was a Macro F$_1$-score of 0.45 on 5-fold cross-validation achieved with the \textit{linear} kernel and $C=10$. The performance of this classifier on the test set is reported in Table \ref{table:c_report}. Its performance is rather bad with a Macro F$_1$-score of 0.38. It has difficulties on predicting cases of medium importance (level 3; $F_1=0.2$) and key cases (level 1; $F_1=0.25$) but has a good recall for cases of low importance (level 4; $R=0.81$) and a decent precision for cases of high importance (level 2; $P=0.67$). It performs better than the majority baseline ($F_1=0.15$) and a random baseline ($F_1=0.32$). We originally expected better results, thus we further investigate these results, the data, and our initial hypotheses.

\begin{table} 
	\centering 
	\begin{tabular}{lrrrr}
		\toprule
		Importance level &  Precision &  Recall &  F1-score &  Support \\
		\midrule
		1            &       0.21 &    0.31 &      0.25 &    13 \\
		2            &       0.67 &    0.29 &      0.40 &    14 \\
		3            &       0.50 &    0.12 &      0.20 &    16 \\
		4            &       0.57 &    0.81 &      0.67 &    32 \\
		\midrule 
		Macro avg    &       0.49 &    0.38 &      0.38 &    75 \\
		\bottomrule
	\end{tabular}
	\caption{Test-set results of predicting case importance using argument-based features.} 
	\label{table:c_report} 
\end{table}

We originally hypothesized that important cases contain more arguments, are longer and more detailed, and also contain rarer argument types, whereas the cases of less importance contain less arguments, are shorter and contain more arguments of the types \textit{Precedents of the ECHR} and \textit{Application to the concrete case}. 

\begin{figure}[h!]
	\centering
	\includegraphics[width=0.65\textwidth]{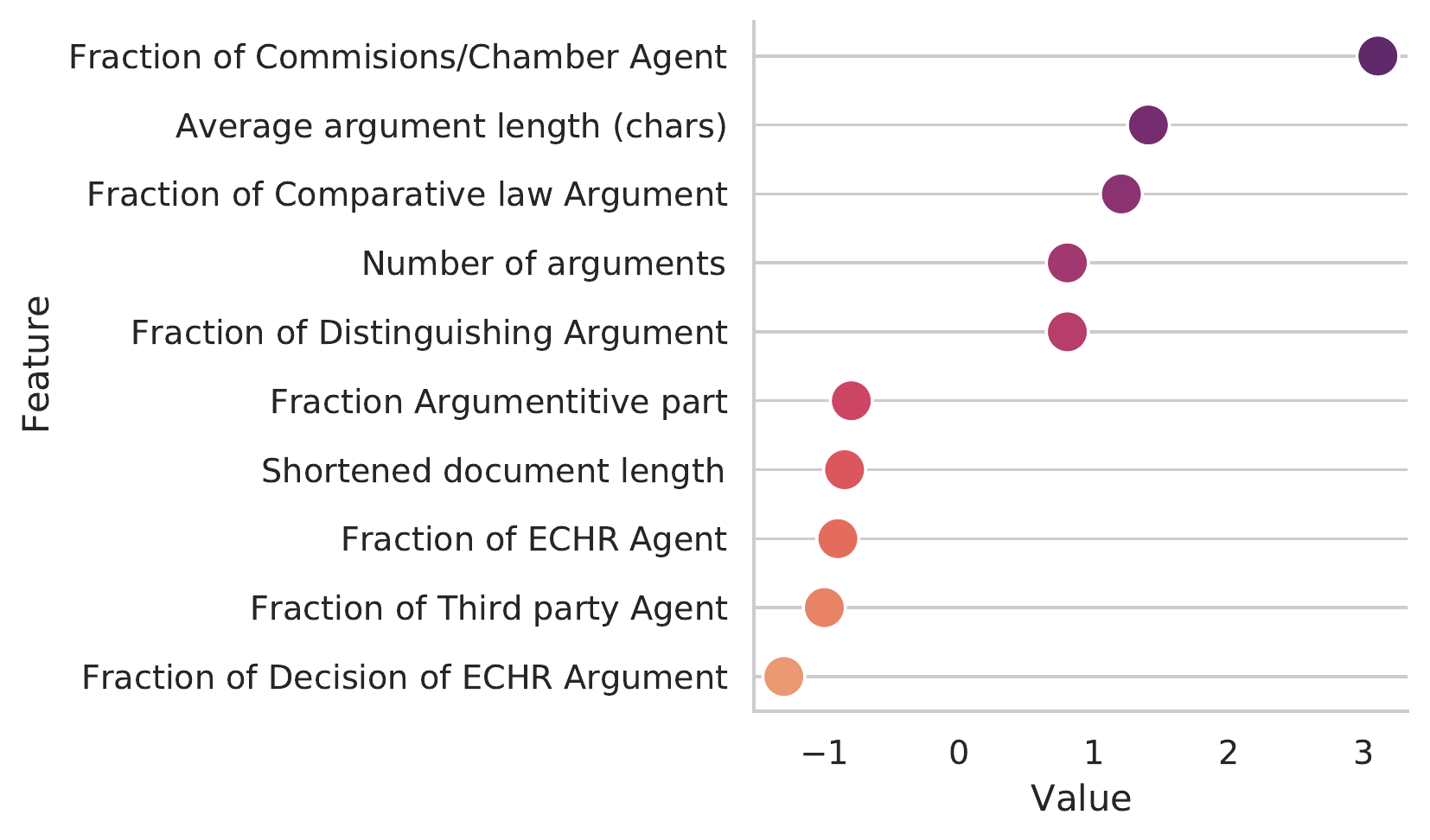}
	\caption{Feature importance of the classifier for importance level 1 vs. level 4.}
	\label{fig:1vs4}
\end{figure}

By examining the importance of individual features from our classifier (see Figure \ref{fig:1vs4}) we can see that especially the fraction of the Commission/Chamber agent is the most indicative when predicting key cases versus cases with low importance. Similarly important are the number of arguments and the average argument length. Furthermore, fraction of the Decision of the ECHR is always in the top five arguments indicative for the class of lower importance, meaning each time the lower importance level has a higher fraction of ECHR Decision arguments. 

Finally, we look into the average value of each feature over all cases (Table \ref{table:avg_importance}) to analyze whether our hypotheses hold. Indeed, the more important a case is, the longer its document length tends to be on average. But we also have to keep in mind that there are a few cases with a length of up to 80,000 words that skew the average document length of importance level 1. When accounting for those by dropping those over 40,000 words long, the average length is still at 14,460 for importance level 1. Surprisingly, also importance level 4 is affected and drops to 7,947. Key cases of level 1 also contain more arguments on average. 

Furthermore, the more important, the longer the average value of the average argument length is. Key cases of importance level 1 also have more arguments on average and contain more rare arguments like \textit{overruling}, \textit{distinguishing}, \textit{margin of appreciation} and \textit{systematic interpretation}. But surprisingly, no clear trend for \textit{Precedents of the ECHR} and \textit{Application to the concrete case} can be seen, so our hypothesis that they should be more common in less important cases seems not to hold. Additionally, \textit{Fraction of Commission/Chamber} is generally more frequent on higher importance levels, except when comparing level 1 and level 2.
However, even though we observed some general trends, it is not possible to reliably distinguish the importance of the case using the argument-based features alone.

\begin{table} 
\centering
	\begin{tabular}{lrrrr}
		\toprule
	&1	&2	&3	&4\\
	\midrule
Doc Length	&17,757	&12,315	&9,906	&8,329\\
Arg Length	&598	&549	&520	&471\\
Shortened Doc Length	&10,571	&6,780	&5,654	&4,862\\
\vspace{0.5em} No. of Args	&52	&36	&38	&39\\
Fraction Argumentative Part	&0.36	&0.32	&0.39	&0.43\\
\vspace{0.5em} Shortened Fraction Argumentative Part	&0.65	&0.63	&0.69	&0.71\\
Fraction of Distinguishing Arg	&0.01	&0.01	&0.00	&0.00\\
Fraction of Margin of Appreciation Arg	&0.01	&0.01	&0.00	&0.00\\
Fraction of Decision ECHR Arg	&0.06	&0.07	&0.09	&0.11\\
Fraction of Non contestation Arg	&0.02	&0.02	&0.02	&0.02\\
Fraction of Overruling Arg	&0.00	&0.00	&0.00	&0.00\\
Fraction of Comparative law Arg	&0.00	&0.00	&0.00	&0.00\\
Fraction of Teleological interpretation Arg	&0.00	&0.01	&0.00	&0.00\\
Fraction of Application case Arg	&0.54	&0.60	&0.51	&0.57\\
Fraction of Systematic interpretation Arg	&0.00	&0.00	&0.00	&0.00\\
Fraction of Necessity/Proportionality Arg	&0.17	&0.10	&0.16	&0.11\\
Fraction of Suitability Arg	&0.00	&0.00	&0.00	&0.00\\
Fraction of Legitimate purpose Arg	&0.02	&0.02	&0.02	&0.02\\
Fraction of Legal basis Arg	&0.04	&0.05	&0.04	&0.04\\
Fraction of Precedents ECHR Arg	&0.13	&0.10	&0.14	&0.13\\
Fraction of Textual interpretation Arg	&0.00	&0.00	&0.00	&0.00\\
Fraction of Applicant Agent	&0.18	&0.18	&0.16	&0.16\\
Fraction of ECHR Agent	&0.61	&0.62	&0.68	&0.68\\
Fraction of State Agent	&0.17	&0.15	&0.16	&0.15\\
Fraction of Commission/Chamber Agent	&0.02	&0.04	&0.00	&0.00\\
Fraction of Third parties Agent	&0.02	&0.01	&0.01	&0.01\\
		\bottomrule
	\end{tabular}
	\caption{Average value over all cases of each feature and each importance level} 
	\label{table:avg_importance} 
\end{table}

%\section{Discussion and limitations} \label{sec:discussion}

\section{Conclusion} \label{chapter:conclusion}

We designed a new annotation scheme for legal arguments in court decisions of the ECHR. Using this scheme, we annotated a large dataset of cases which we make available to the research community under open license. We built a robust argument mining model that can function with the particular challenges of the legal text in the ECHR decisions. We have shown that the bigger, more powerful general model \robalg\ can compete with the smaller, to the general legal domain adapted \lbert. 

This motivated us to examine different domain adaption approaches, namely further pretraining of an existing language model on European legal data. This led to significant performance improvements of up to 11\% for the argument type prediction.

We evaluated our best model in detail where we found that in general the model is able to predict classes quite well if sufficient support for it exists in the dataset, whereas for the really rare classes with nearly any support it failed to predict them. There were some exceptions to the general trend for which we thoroughly analyzed the errors by cooperating with legal experts. In these cases we found that there is often not enough context available for predicting an argument. But without more context our model predictions often seemed more convincing.

Furthermore, we examined the models ability to generalize over decisions from different ECHR Articles. We have reported a up to 27\% decreased performance, indicating that decisions concerning different Articles can differ significantly.

Finally, we utilized the arguments of a decision for predicting its importance level. To this end, we built an SVM model which allowed for introspection of the argument importance. This model was unable to discriminate well and by looking further into it, we saw that, despite showing slight tendencies for some features, most of the ECHR decisions have similar features. We conclude from this that the ECHR works diligently and holds itself to the same high standards regardless of the importance level of a case.

\section*{Acknowledgments}
This project was supported by RMU-Initiativsfond Forschung (Förderlinie 2), by the DFG project ECALP (HA 8018/2-1), and by the German Federal Ministry of Education and Research and the Hessen State Ministry for Higher  Education, Research and the Arts within their joint support of the National Research Center for Applied Cybersecurity ATHENE (research area ``Legal Aspects of Privacy and IT Security''). The independent research group TrustHLT is supported by the Hessian Ministry of Higher Education, Research, Science and the Arts. We are extremely thankful to our team of annotators, namely 
Isabelle Baumann, Gaspar Fink, Nele Musch, Joice Roberts, Sina Schmitt, and Diya Varghese. Thanks to Fabian Kaiser for designing the initial pilot studies. Special thanks to Richard Eckart de Castilho and Jan-Christoph Klie for their tireless support with the INCEpTION annotation platform and to Leonard Bongard for implementing the agreement studies.

\bibliography{bibliography}

\begin{appendices}

\section{ECHR judgment example}
\label{appendix:judgment}

\begin{footnotesize}
CASE OF BERREHAB v. THE NETHERLANDS

(Application no. 10730/84)

JUDGMENT

STRASBOURG 

21 June 1988

In the Berrehab case[*],
The European Court of Human Rights, sitting, in accordance with Article 43 (art. 43) of the Convention for the Protection of Human Rights and Fundamental Freedoms ("the Convention") and the relevant provisions of the Rules of Court, as a Chamber composed of the following judges:
Mr. R. Ryssdal, President,
Mr. Thór Vilhjálmsson,
...,
and also of Mr. M.-A. Eissen, Registrar, and Mr. H. Petzold, Deputy Registrar,

Having deliberated in private on 26 February and 28 May 1988,

Delivers the following judgment, which was adopted on the last-mentioned date:

\textbf{PROCEDURE}

1.   The case was referred to the Court by the European Commission of Human Rights ("the Commission") and by the Netherlands Government ("the Government") on 13 March and 10 April 1987 respectively, within the three-month period laid down in Article 32 § 1 and Article 47 (art. 32-1, art. 47) of the Convention. It originated in an application (no. 10730/84) against the Kingdom of the Netherlands lodged with the Commission under Article 25 (art. 25) by a Moroccan national, Abdellah Berrehab, a Netherlands national, Sonja Koster, and their daughter Rebecca Berrehab, likewise of Netherlands nationality, on 14 November 1983. "The applicants" hereinafter means only Abdellah and Rebecca Berrehab, as the Commission declared Sonja Koster’s complaints inadmissible (see paragraph 18 below).
...

\textbf{AS TO THE FACTS}

I.   THE CIRCUMSTANCES OF THE CASE

7.   Mr. Berrehab, a Moroccan citizen born in Morocco in 1952, was permanently resident in Amsterdam at the time when he applied to the Commission.
His daughter Rebecca, who was born in Amsterdam on 22 August 1979, has Netherlands nationality. She is represented by her guardian, viz. her mother, Mrs. Koster, who is likewise a Netherlands national.
...

II.  THE RELEVANT LEGISLATION, PRACTICE AND CASE-LAW

A. The general context of Netherlands immigration policy
...

PROCEEDINGS BEFORE THE COMMISSION

17.  In their application of 14 November 1983 to the Commission (no. 10730/84), Mr. Berrehab and his ex-wife Mrs. Koster, the latter acting in her own name and as guardian of their under-age daughter Rebecca, alleged that Mr. Berrehab’s deportation amounted - in respect of each of them, and more particularly for the daughter - to treatment that was inhuman and therefore contrary to Article 3 (art. 3) of the Convention. In their submission, the deportation was also an unjustified infringement of the right to respect for their private and family life, as guaranteed in Article 8 (art. 8).
...

\textbf{AS TO THE LAW}

I.   ALLEGED VIOLATION OF ARTICLE 8 (art. 8)

19.  In the applicants’ submission, the refusal to grant a new residence permit after the divorce and the resulting expulsion order infringed Article 8 (art. 8) of the Convention, which provides:
...

A. Applicability of Article 8 (art. 8)

20.  The applicants asserted that the applicability of Article 8 (art. 8) in respect of the words "right to respect for ... private and family life" did not presuppose permanent cohabitation. The exercise of a father’s right of access to his child and his contributing to the cost of education were also factors sufficient to constitute family life.
...

B. Compliance with Article 8 (art. 8)
...

II.  ALLEGED VIOLATION OF ARTICLE 3 (art. 3)
...

III.  APPLICATION OF ARTICLE 50 (art. 50)

32.  By Article 50 (art. 50) of the Convention,
...

34.  The Court shares the view of the Commission. Taking its decision on an equitable basis, as required by Article 50 (art. 50), it awards the applicants the sum of 20,000 guilders.

\textbf{FOR THESE REASONS, THE COURT}

1. Holds by six votes to one that there has been a violation of Article 8 (art. 8);

2. Holds unanimously that there has been no violation of Article 3 (art. 3);

3. Holds unanimously that the Netherlands is to pay to the applicants 20,000 (twenty thousand) Dutch guilders by way of just satisfaction;

4. Rejects unanimously the remainder of the claim for just satisfaction.

Done in English and in French, and delivered at a public hearing in the Human Rights Building, Strasbourg, on 21 June 1988.

Rolv RYSSDAL,
President

Marc-André EISSEN,
Registrar
\end{footnotesize}

\section{Argument Types Examples} \label{appendix:arg_types}

\paragraph{Procedural arguments -- Non contestation by the parties}

It is uncontested that the administrative conviction had become “final” before the criminal proceedings began in respect of the first applicant. That fact was acknowledged by the courts acting in the criminal case and reflected in the judgment of 18 August 2014.

\paragraph{Method of interpretation -- Textual interpretation}

Were this is not the case, it would not have been necessary to add the word “punished” to the word \emph{tried} since this would be mere duplication. Article 4 of Protocol No. 7 applies even where the individual has merely been prosecuted in proceedings that have not resulted in a conviction.

\paragraph{Method of interpretation -- Historical interpretation}
	
The Court also notes that, apart from revealing that the non bis in idem rule was construed relatively narrowly, the travaux préparatoires on Protocol No. 7 shed little light on the matter.

\paragraph{Method of interpretation -- Systematic interpretation}

The ne bis in idem principle is mainly concerned with due process, which is the object of Article 6, and is less concerned with the substance of the criminal law than Article 7. The Court finds it more appropriate, for the consistency of interpretation of the Convention taken as a whole, for the applicability of the principle to be governed by the same, more precise criteria as in Engel.

\paragraph{Method of interpretation -- Teleological interpretation}
	
The object of Article 4 of Protocol No. 7 is to prevent the injustice of a person’s being prosecuted or punished twice for the same criminalised conduct.

\paragraph{Method of interpretation -- Comparative law}

% \todo{Better example}
Four States (Germany, the Netherlands, Turkey and the United Kingdom) have not ratified the Protocol; and one of these (Germany) plus four States which did ratify (Austria, France, Italy and Portugal) have expressed reservations or interpretative declarations to the effect that \emph{criminal} ought to be applied to these States in the way it was understood under their respective national laws.

\paragraph{Test of the principle of proportionality -- Legal basis}
	
To be justified under Article 10 § 2 of the Convention, an interference with the right to freedom of expression must have been say{prescribed by law}

\paragraph{Test of the principle of proportionality -- Legitimate purpose}
	
The aim of the Swedish personnel control system is clearly a legitimate one for the purposes of Article 8 (art. 8), namely the protection of national security.

\paragraph{Test of the principle of proportionality -- Suitability}

The broader concept of proportionality inherent in the phrase ``necessary in a democratic society'' requires a rational connection between the measures taken by the authorities and the aim that they sought to realise through these measures, in the sense that the measures were reasonably capable of producing the desired result.

\paragraph{Test of the principle of proportionality -- Necessity/Proportionality}
%\todo{better example}
Fair balance to be struck between duly safeguarding the interests of the individual protected by the ne bis in idem principle, on the one hand, and accommodating the particular interest of the community in being able to take a calibrated regulatory approach in the area concerned, on the other.

\paragraph{Institutional arguments -- Overruling}

In the light of the foregoing considerations, the Court takes the view that it is necessary to depart from the case-law established by the Commission in the case of X v. Germany and affirm that Article 7 § 1 of the Convention guarantees not only the principle of non-retrospectiveness of more stringent criminal laws but also, and implicitly, the principle of retrospectiveness of the more lenient criminal law.

\paragraph{Institutional arguments -- Distinguishing}

It therefore remains only to ascertain whether the provisions (art. 5, art. 64) applied in the present case are covered by that reservation. They differ in certain essential respects from those in issue in the Chorherr case.

\paragraph{Institutional arguments -- Margin of Appreciation}
	
It is in the first place for the Contracting States to choose how to organise their legal system, including their criminal-justice procedures.

\paragraph{Precedents of the ECHR}

The notion of the ``same offence'' – the idem element of the ne bis in idem principle in Article 4 of Protocol No. 7 – is to be understood as prohibiting the prosecution or conviction of a second ``offence'' in so far as it arises from identical facts or facts which are substantially the same (see Sergey Zolutukhin, cited above, §§ 78-84).

\paragraph{Decision of the ECHR}

The Court considers that this complaint is not manifestly ill founded within the meaning of Article 35 § 3 of the Convention or inadmissible on any other grounds. It must therefore be declared admissible.

\paragraph{Application to the concrete case}

The Court observes that the first applicant was convicted of perpetrating acts of mass disorder, which had contributed to disruption of the assembly. The attribution of responsibility for those acts was therefore a central question in the determination of the criminal charges against him. In those circumstances, his complaint about the authorities’ role in the occurrence of the disorder is inseparable from his complaint concerning the lack of justification for his criminal liability. For that reason, the Court is not required to assess, as a separate issue under Article 11 of the Convention, the authorities’ alleged failure to discharge their positive obligation in respect of the conduct of the demonstration at Bolotnaya Square.

\section{Supplementary data analysis}
\label{sec:bio-tag-distribution}

See tables \ref{table:bio_arg_stats}, \ref{table:bio_ag_stats}, \ref{table:train_arg_stats}, \ref{table:train_ag_stats}, \ref{table:test_arg_stats}, and \ref{table:test_ag_stats} which are also referenced from the main body.

\begin{table} 
	\centering 
	\begin{tabular}{ lrr } 
		\toprule
		Argument Type BIO Labels & Frequency & $\approx$ in \% \\ 
		\midrule
		O & 867,913 & 36.24 \\
		I-Application case & 820,866 & 34.27 \\
		I-Precedents ECHR & 309,316 & 12.91 \\
		\vspace{0.5em} I-Necessity/Proportionality & 258,767 & 10.80 \\
		I-Legal basis & 41,424 & 1.73 \\
		I-Decision ECHR & 36,260 & 1.51 \\
		\vspace{0.5em} I-Legitimate purpose & 13,445 & 0.56 \\
		I-Non contestation & 9,743 & 0.41 \\
		B-Application case & 8,688 & 0.36 \\
		I-Distinguishing & 7,929 & 0.33 \\
		I-Margin of Appreciation & 5,308 & 0.22 \\
		I-Teleological interpretation & 4,250 & 0.18 \\
		I-Suitability & 2,131 & 0.09 \\
		B-Precedents ECHR & 2,015 & 0.08 \\
		B-Necessity/Proportionality & 1,958 & 0.08 \\
		B-Decision ECHR & 1,236 & 0.05 \\
		I-Comparative law & 1,110 & 0.05 \\
		\vspace{0.5em} I-Textual interpretation & 1,053 & 0.04 \\
		B-Legal basis & 513 & 0.02 \\
		B-Legitimate purpose & 271 & 0.01 \\
		I-Systematic interpretation & 271 & 0.01 \\
		B-Non contestation & 264 & 0.01 \\
		\vspace{0.5em} I-Overruling & 109 & 0.00 \\
		B-Distinguishing & 88 & 0.00 \\
		B-Margin of Appreciation & 67 & 0.00 \\
		B-Teleological interpretation & 50 & 0.00 \\
		B-Suitability & 25 & 0.00 \\
		B-Comparative law & 13 & 0.00 \\
		\vspace{0.5em} B-Textual interpretation & 10 & 0.00 \\
		B-Systematic interpretation & 5 & 0.00 \\
		B-Overruling & 2 & 0.00 \\
		\bottomrule
	\end{tabular} 
	\caption{Token-level statistics of each token's argument type label (using BIO tagging) in the entire annotated corpus}
	\label{table:bio_arg_stats} 
\end{table}

\begin{table} 
	\centering 
	\begin{tabular}{ lrr } 
		\toprule
		Agent Type BIO Label & Frequency & in \% \\ 
		\midrule
		I-ECHR & 1,053,629 & 43.99 \\
		O & 867,913 & 36.24 \\
		I-State & 220,293 & 9.20 \\
		I-Applicant & 201,605 & 8.42 \\
		I-Third parties & 22,431 & 0.94 \\
		I-Commission/Chamber & 14,024 & 0.59 \\
		B-ECHR & 9,950 & 0.42 \\
		B-Applicant & 2,471 & 0.10 \\
		B-State & 2,399 & 0.10 \\
		B-Third parties & 212 & 0.01 \\
		B-Commission/Chamber & 173 & 0.01 \\
		\bottomrule
	\end{tabular} 
	\caption{Token-level statistics of each token's agent label (using BIO tagging) in the entire annotated corpus} 
	\label{table:bio_ag_stats} 
\end{table}

\begin{table} 
	\centering 
	\begin{tabular}{ lrr } 
		\toprule
		Argument Type & Frequency & in \% \\ 
		\midrule
		Application to the concrete case & 7,031 & 57.44 \\
		Precedents of the ECHR & 1,608 & 13.14 \\
		Test of the principle of proportionality -- Proportionality & 1,597 & 13.05 \\
		Decision of the ECHR & 977 & 7.98 \\
		Test of the principle of proportionality -- Legal basis & 400 & 3.27 \\
		Test of the principle of proportionality -- Legitimate purpose & 224 & 1.83 \\
		Non contestation by the parties & 216 & 1.76 \\
		Distinguishing & 66 & 0.54 \\
		Margin of Appreciation & 45 & 0.37 \\
		Teleological interpretation & 33 & 0.27 \\
		Test of the principle of proportionality -- Suitability & 23 & 0.19 \\
		Comparative law & 10 & 0.08 \\
		Textual interpretation & 7 & 0.06 \\
		Systematic interpretation & 3 & 0.02 \\
		Overruling & 1 & 0.01 \\
		\bottomrule
	\end{tabular} 
	\caption{Number of text spans associated with a particular argument type in the training data} 
	\label{table:train_arg_stats} 
\end{table}

\begin{table} 
	\centering 
	\begin{tabular}{ lrr } 
		\toprule 
		Agent & Frequency & in \% \\ 
		\midrule
		ECHR & 8,036 & 65.65 \\
		Applicant & 2,001 & 16.35 \\
		State & 1,933 & 15.79 \\
		Third parties & 142 & 1.16 \\
		Commission/Chamber & 129 & 1.05 \\
		\bottomrule
	\end{tabular} 
	\caption{Number of text spans associated with a particular argument agent in the training data} 
	\label{table:train_ag_stats} 
\end{table}

\begin{table} 
	\centering 
	\begin{tabular}{ lrr } 
		\toprule
		Argument Type & Frequency & in \% \\ 
		\midrule
		Application to the concrete case & 801 & 56.17 \\
		Precedents of the ECHR & 170 & 11.92 \\
		Test of the principle of proportionality -- Proportionality & 169 & 11.85 \\
		Decision of the ECHR & 129 & 9.05 \\
		Test of the principle of proportionality -- Legal basis & 65 & 4.56 \\
		Non contestation by the parties & 28 & 1.96 \\
		Test of the principle of proportionality -- Legitimate purpose & 18 & 1.26 \\
		Distinguishing & 16 & 1.12 \\
		Margin of Appreciation & 12 & 0.84 \\
		Teleological interpretation & 12 & 0.84 \\
		Comparative law & 2 & 0.14 \\
		Systematic interpretation & 1 & 0.07 \\
		Textual interpretation & 1 & 0.07 \\
		Overruling & 1 & 0.07 \\
		Test of the principle of proportionality -- Suitability & 1 & 0.07 \\
		\bottomrule
	\end{tabular} 
	\caption{Number of text spans associated with a particular argument type in the test data} 
	\label{table:test_arg_stats} 
\end{table}

\begin{table} 
	\centering 
	\begin{tabular}{ lrr } 
		\toprule
		Agent & Frequency & in \% \\ 
		\midrule
		ECHR & 915 & 64.17 \\
		Applicant & 238 & 16.69 \\
		State & 219 & 15.36 \\
		Third parties & 28 & 1.96 \\
		Commission/Chamber & 26 & 1.82 \\
		\bottomrule
	\end{tabular} 
	\caption{Number of text spans associated with a particular argument agent in the test data} 
	\label{table:test_ag_stats} 
\end{table}

\section{Failed pretraining from scratch} \label{appendix:scratch}

We also extensively experimented with pre-training a model from scratch using solely legal data and legal word-piece vocabulary. Despite our efforts and intensive utilization of GPU resources, this experiment failed. Yet we believe that sharing the experiment details might be helpful for future work.

We pre-train a model from scratch similarly to \lbert\ \cite{Chalkidis.et.al.2020.Findings}. As \citet{Gu.et.al.2022.ACM.CH} have shown improvements using their own vocabulary and whole-word masking (WWM) in the biomedical domain, we also train our own tokenizer to have an in-domain vocabulary and utilize whole-word masking. WWM improves the normal masking process in that it masks out the whole original word if any subword of it is masked.

We see legal court cases as the domain of our interest, thus in contrast to \lbert\ we only use court cases as our data. Since we need a lot of data for pre-raining from scratch, we additionally use the \claw\ dataset which consists of 4,938,129 US Court cases, amounting to 43.2 GB, along to our ECHR dataset which consists of 65,908 English cases, amounting to 1.1 GB. Because we are also interested in the effect of different legal systems, we first pretrain our model only on the \claw\ dataset for the majority of our steps, save the checkpoint and then further pretrain it only on the ECHR dataset. That way, we have two models, one pretrained on the \claw\ dataset and one combined model with finetuning to our ECHR use case. With these we can study possible differences of our model in regard to the training on Common Law, where priority is given to jurisprudence over doctrine, and Civil Law, where the opposite is true \cite{Mullerat.2008}. 

We use both datasets for the training of our legal court tokenizer. The tokenizer we use is a BERT WordPiece tokenizer with a vocabulary size of 50,000, the size of the RoBERTa tokenizer.
An advantage of using our own in-domain vocabulary is that the input will be shorter, which should make it easier for the model to learn. Additionally, it can enable our model to learn a meaningful single specific representation for specific legal words. A comparison of this can be seen in Table \ref{table:tok_comp}.

\begin{table}
	\centering
	\begin{tabular}{ lrrr } 
		\toprule
		& \lbert\ & \roba\ & Our model \\ 
		\midrule
		Average document length & 7,078 & 8,696 & 7,022 \\ 
		Average input length & 68 & 83 & 67 \\ 
		Example: \texttt{Affidavit} & \texttt{affidavit} & \texttt{Aff}, \texttt{idav}, \texttt{id} & \texttt{Affidavit} \\ 
		\bottomrule
	\end{tabular}
	\caption{Comparison of the effect of different tokenization. \texttt{Affidavit} is a statement made under an oath.}
	\label{table:tok_comp}
\end{table}

We pretrain our model with the same architecture and hyperparameters as \robalg\ and a batch size of 2,064. Originally, we planned to pretrain on the \claw\ corpus for 90,000 steps and then on the ECHR corpus for 10,000--20,000 steps. Unfortunately, due to huge computational costs and limited resources, we could only pretrain for 20,000 and 5,000 steps, respectively.

During pre-training, both the train and validation loss stagnated after around 5,000 steps. This is surprising because it contradicts other findings reporting a better performance after pretraining from scratch \citep{Gu.et.al.2022.ACM.CH,Zheng.et.al.2021.ICAIL,Chalkidis.et.al.2020.Findings}.  Possible reasons for our stagnation here could be that our data is not diverse enough or we encountered bad luck by our model initialization which led to a dead end. The beginning of the stagnation at around 5k steps also might suggest that our learning rate was too big because we first use a warm-up for 5,400 steps. Warm-up means the learning rate linearly increases from 0 until it reaches the specified learning rate for the model at the end of the warm-up steps. We suspect that we either need more warm-up steps until the learning stabilizes or directly select a smaller learning rate. 

\end{appendices}
	
\end{document}